\pdfoutput=1
\documentclass{article} 
\usepackage[preprint]{neurips_2026}
\setcitestyle{round}
\usepackage[utf8]{inputenc} 
\usepackage[T1]{fontenc}    
\usepackage{url}            
\usepackage{booktabs}       
\usepackage{amsfonts}       
\usepackage{nicefrac}       
\usepackage{microtype}      
\usepackage{xcolor}         

\usepackage{tabularx}
\usepackage{array}

\usepackage{algorithm}
\usepackage{algpseudocode}
\usepackage{hyperref}

\algrenewcommand\algorithmicrequire{\textbf{Input:}}
\algrenewcommand\algorithmicensure{\textbf{Output:}}

\usepackage{amsmath,amsfonts,bm}

\def\eqref#1{equation~\ref{#1}}

\def\1{\bm{1}}

\DeclareMathAlphabet{\mathsfit}{\encodingdefault}{\sfdefault}{m}{sl}
\SetMathAlphabet{\mathsfit}{bold}{\encodingdefault}{\sfdefault}{bx}{n}

\DeclareMathOperator*{\argmin}{arg\,min}

\usepackage{graphicx}
\usepackage{wrapfig}
\usepackage{enumitem}

\usepackage{titlesec}

\title{Optimal Transport Meets t-SNE: Integrated Vis\-ualization without Batch Effects in \mbox{OT-tSNE}}

\author{Azhar Akhmetova\thanks{Equal contribution.} \\
University of T\"{u}bingen \\
\texttt{azhar.akhmetova}\\\texttt{@uni-tuebingen.de} \\
\And
Edward Eichhorn\footnotemark[1]\\
University of T\"{u}bingen \\
\texttt{edward.eichhorn}\\\texttt{@student.uni-tuebingen.de} \\
\And
Boris Landa \\
Yale University\\
\texttt{boris.landa@yale.edu} \\
\And
Dmitry Kobak \\
Ghent University and VIB.AI\\
\texttt{dmitry.kobak@ugent.be} \\
\And
Sebastian Damrich\\
University of T\"{u}bingen \\
\texttt{sebastian.damrich@uni-tuebingen.de} \\
}

\begin{document}

\maketitle

\begin{abstract}
Visual exploration is an integral step in understanding the structure of datasets. However, systematic variation between datasets collected under different conditions, known as batch effects, can obscure shared structure and complicate their joint visualization. Existing methods often rely on cross-batch nearest-neighbor relations to integrate the batches. Yet, these local correspondences can be unreliable for challenging cross-batch geometries. We propose optimal-transport \mbox{t-SNE}~(OT-tSNE), which replaces nearest-neighbor-based matching with globally coupled affinities derived from entropic optimal transport. Controlled synthetic experiments demonstrated that nearest-neighbor-based methods can deteriorate, while optimal-transport affinities and OT-tSNE embeddings remain robust. Across seven real single-cell datasets, we benchmarked OT-tSNE against a broad range of integration and visualization methods and found that it achieves a Pareto-efficient trade-off between batch mixing and biological conservation, and integrates batches particularly well.
\end{abstract}

\section{Introduction}
\label{sec:intro}

As many scientific disciplines become more and more data driven, the need for large datasets increases. However, acquiring several smaller datasets is often more feasible than a single large one. Unfortunately, separately recorded datasets regularly show structural differences. Such differences can be technical nuisances like varying measurement techniques or protocols employed, or signals like donors or donor sex, that are undesirable for the task at hand. We collectively call such structural differences \textit{batch effect} and the different smaller datasets \textit{batches}. Many methods have been proposed to remove the batch effect and to integrate batches~\citep{luecken2022benchmarking}. These typically integrate the data in the original feature space or in a relatively high-dimensional latent space. 

As a result, these general-purpose batch removal methods are not directly usable when the aim is to visually explore the data. Visualizing the data as 2D scatter plots, e.g., with neighbor embedding methods like t-SNE~\citep{maaten2008visualizing}, has become an important step in many exploratory data analysis pipelines~\citep{debodt2025lowdim}. General-purpose batch integration excels at producing good representations for downstream analysis, which then can be visualized with tools like t-SNE. But when visualization is the primary aim, directly producing an integrated visualization is both conceptually simpler and can yield better results. 
\begin{figure}
    \centering
    \includegraphics[width=\linewidth]{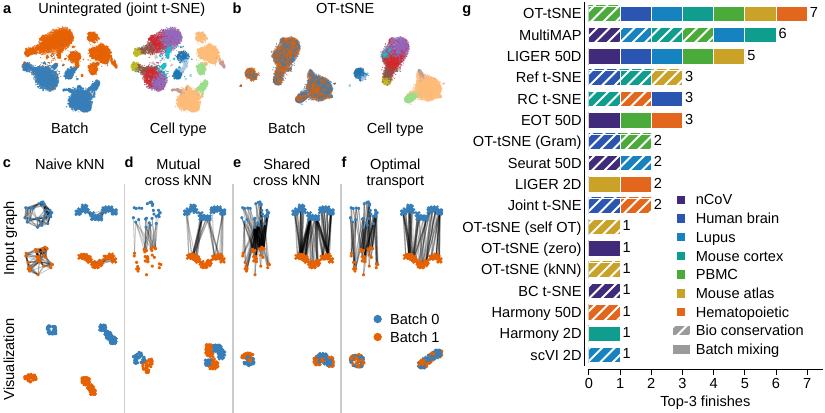}
    \caption{\textbf{a.} Unintegrated t-SNE plot of the Lupus~\citep{kang2018multiplexed} data shows a clear batch effect. \textbf{b.} Our OT-tSNE integrates the batches. \textbf{c -- f.} t-SNE with various affinity graphs on 2D toy data. \mbox{\textbf{d -- f}} use the cross-batch edges in addition to the kNN graph. The cross-kNN variants only move the clusters close to each other (``kissing problem''), while optimal transport integrates them. \textbf{g.} OT-tSNE and MultiMAP achieved most top-3 ranks across all evaluation metrics and datasets.}
    \label{fig:fig1}
    \vspace*{-1mm}
\end{figure}
For this goal, we propose \textit{optimal-transport t-SNE (OT-tSNE)}, a novel approach combining optimal transport~\citep{gabriel2019computational} and the popular t-SNE algorithm. Vanilla t-SNE lays the k-nearest-neighbor graph (kNN graph) out in 2D. In the presence of a strong batch effect, no kNN edge crosses from one batch to the other, leading to fully separated batches in the visualizations (Fig.~\ref{fig:fig1}a, c). The key idea of OT-tSNE is to augment this graph with explicit cross-batch edges before computing the layout. Many existing tools rely on cross-batch k-nearest neighbors, either mutual nearest neighbors, when both points are each other's nearest neighbors~\citep{haghverdi2018mnn, stuart2019seuratcca, hie2019scanorama, barkas2019conos} or shared nearest neighbors, when at least one point is the nearest neighbor of the other~\citep{polanski2020bbknn, jain2021multimap}. Nearest neighbors are inherently local as each point selects nearest neighbors separately. This can lead to hubs among the cross-batch edges, especially on the fringes of the batches~\citep[``kissing problem'']{fastMNN}, which can deteriorate the integration in the visualization (Fig.~\ref{fig:fig1}d,e). Instead, optimal transport computes the overall best association between both batches, while ensuring that each point contributes equally. This allows OT-tSNE to produce very well-integrated visualizations while retaining the structure of individual batches (Fig.~\ref{fig:fig1}b,f). 

We corroborate the strong performance of OT-tSNE with a large benchmark comparing 23 method configurations for visualizing data with batch effect in 2D across 7 datasets. 
We found that OT-tSNE matches the state-of-the-art combination of batch mixing and preservation of data structure, particularly outperforming the competitors in terms of batch integration (Fig.~\ref{fig:fig1}g).

In sum, our contributions are
\begin{enumerate}[nosep, leftmargin=0.5\leftmargin]
    \item the first combination of optimal transport and neighbor embedding methods, OT-tSNE, for integrated visualizations of data with batch effect,
    \item illustrating that nearest-neighbor-based methods can struggle even if the batch effect is orthogonal to the data, while OT-tSNE remains unaffected,
    \item an extensive benchmark for visualizing batched data, highlighting the strong performance of OT-tSNE especially for batch mixing,
    \item an open-source implementation of OT-tSNE based on the \texttt{pot} and \texttt{opentsne} packages.
\end{enumerate}

Our code is available at \url{https://github.com/NDR-lab/ot-tsne}.

\section{Related work}
\label{sec:related_work}
Our work draws on two main strands of research, visualization and optimal transport. 
Spectral visualization approaches, such as Laplacian Eigenmaps~\citep{belkin2003laplacian} or Diffusion Maps~\citep{coifman2006diffusion}, lay out a graph approximation of the data via the eigenvectors of the graph's Laplacian. Neighbor embeddings, like t-SNE~\citep{maaten2008visualizing} and UMAP~\citep{mcinnes2018umap}, often produce visualizations with finer resolution and have thus become the de facto standard for non-linear visualization. Such embeddings distort distances between observations and fail to capture global data geometry~\citep{chari2023specious}. However, when the high-dimensional data are well clustered, the groups visible in t-SNE plots often agree with those identified by dedicated clustering methods~\citep{kobak2019art}.

Optimal transport (OT)~\citep{gabriel2019computational} computes how to transport mass from one dataset to another at minimal cost. It became popular in machine learning with \citet{cuturi2013sinkhorn}'s seminal paper on the fast computation of entropically regularized optimal transport with the Sinkhorn-Knopp algorithm. SNEkhorn~\citep{assel2023snekhorn} used optimal transport maps for visualization by replacing the neighbor-based t-SNE affinities with optimal transport plans. This has some resemblance to our proposed OT-tSNE, but does not address batch integration.

The ability of optimal transport to meaningfully relate different datasets makes it well-suited to align datasets or remove batch effects. \citet{courty2016optimal} mapped images from a source domain to the average of images in a target domain, weighted by the transport plan. Similarly, \citet{demetci2022scot} used optimal transport to align multimodal single-cell data. \citet{landa2024entropic}'s entropic OT eigenmaps use an entropic OT (EOT) plan to align two datasets, thus extending Laplacian Eigenmaps and Diffusion Maps to data alignment. Like Laplacian Eigenmaps and Diffusion Maps, it struggles to compress rich information into the very low dimensionality needed for visualization.

Batch correction and integration are major topics in the analysis of single-cell RNA-sequencing (scRNA-seq) data~\citep{luecken2022benchmarking}. Methods like scVI~\citep{lopez2018scvi} and LIGER~\citep{welch2019liger} learn to reconstruct cells from batch-aware encoding models. Most batch correction approaches infer cross-batch associations, often in a dimensionality-reduced space. Harmony~\citep{korsunsky2019harmony} tries to infer compact cross-batch clusters. MNN~\citep{haghverdi2018mnn}, Seurat~\citep{stuart2019seuratcca}, Scanorama~\citep{hie2019scanorama} and others leverage mutual nearest neighbors (MNNs) across batches to estimate local batch-effect vectors. A few very recent methods use optimal transport as part of their batch integration pipeline to determine the integration order for multiple batches~\citep{houdjedj2025scituna}, to regularize adversarial integration~\citep{jiang2026iDLC}, or as pre-alignment before the main contrastive VAE integration step~\citep{zhang2026scBIOT}. None of them uses optimal transport as the main alignment ingredient, or produces 2D visualizations. 

Some batch correction methods output a graph containing both within-batch affinities and cross-batch affinities. These methods are similar in spirit to ours, especially since the resulting graphs are often visualized with UMAP. However, their cross-batch edges are kNN-~\citep[BBKNN]{polanski2020bbknn} or MNN-based~\citep[CONOS]{barkas2019conos} and can thus lead to imbalanced cross-batch affinities and the kissing problem~\citep{fastMNN}.

Closest to our work are approaches that directly produce an integrated 2D visualization. \citet{policar2023embedding} propose to use the t-SNE embedding of one batch as a reference for the other batch. Compound-SNE~\citep{cess2024compoundSNE} relies on knowledge of shared cell types across different batches and alternates Procrustes alignment to a reference with t-SNE embedding optimization.

Other methods are symmetric in their treatment of batches. Batch-corrected t-SNE~\citep[BC t-SNE]{aliverti2020projected} first linearly integrates the input data to make it uncorrelated with the nuisance label and then restricts the t-SNE gradient updates to be orthogonal to the nuisance label. Conditional t-SNE~\citep{kang2021conditional} incorporates the batch information into t-SNE's embedding affinities so that points from different batches move closer together. Revised conditional t-SNE~\citep[RC t-SNE]{heiter2023rctsne} instead changes the input affinities to include inter-batch affinities. MultiMAP~\citep{jain2021multimap} extends UMAP to a multi-dataset visualization method, equipping a graph similar to the BBKNN graph with UMAP-like weights. Deep Visualization~\citep{xu2023structure} trains a neural network to produce visualizations with a UMAP-like loss and uses the batch information in the embedding affinities, similar to conditional t-SNE. 
Our method OT-tSNE also adapts t-SNE for batch-corrected visualizations, but relies on optimal transport to provide cross-batch affinities.

\section{Method}
\label{sec:method}

\begin{figure}[tb]
    \centering\includegraphics[width=\linewidth]{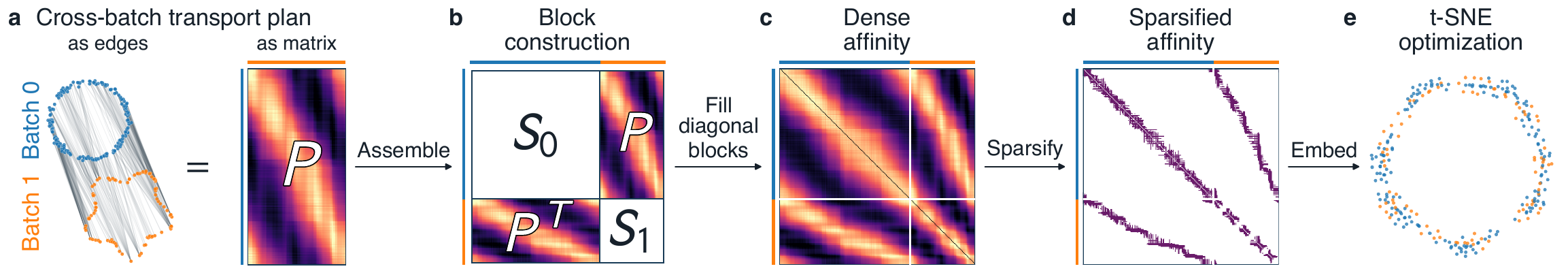}
    \caption{OT-tSNE method overview. \textbf{a.} Toy dataset with batch effect and selected optimal transport cross-batch affinities shown as edges (left), with the full plan shown as matrix $P$ (right). \textbf{b.} Construction of the four-block joint affinity matrix for the t-SNE backend. The plans $P,P^\top$ form the off-diagonal blocks. \textbf{c.}~Completing the on-diagonal blocks $S_0$ and $S_1$ to reflect similarity of points in the same batch. \textbf{d.}~Batch-size-aware sparsification of the dense joint affinities to make them suitable for t-SNE optimization. \textbf{e.}~Integrated OT-tSNE embedding obtained via t-SNE optimization.}
    \label{fig:method}
\end{figure}

\textbf{t-SNE background.}
t-SNE constructs a low-dimensional representation $Y=(y_1, \dots, y_n)^\top\in\mathbb{R}^{n\times d'}$ of a high-dimensional dataset $X\in\mathbb{R}^{n\times d}$, with $d'\ll d$. It does so by aligning pairwise similarity distributions in the high- and low-dimensional spaces~\citep{maaten2008visualizing}. The high-dimensional similarities $p_{ij}$ are symmetrized Gaussian affinities with locally adapted bandwidths, whereas the low-dimensional similarities are defined using a normalized Cauchy kernel
$$q_{ij} = (1+\|y_i - y_j\|^2)^{-1}/Z, \qquad Z = \sum_{kl} (1+\|y_k - y_l\|^2)^{-1}.$$
Previous work has shown that replacing the Gaussian affinities with a binary shared $k$-nearest-neighbor graph produces very similar t-SNE embeddings~\citep{artemenkov2020ncvis, boehm2022attraction,damrich2023from}.
To align the similarities and thus compute the embedding coordinates $Y$, t-SNE minimizes the Kullback-Leibler divergence between $q$ and $p$
\begin{equation}
    \mathcal{L}(Y) = \text{KL}\big(p \| q(Y)\big) = \sum_{ij}p_{ij}\log(p_{ij}/q_{ij}).
\end{equation}
\textbf{OT-tSNE.}
Let $X_b=(x_1^{(b)}, \dots, x_{n_b}^{(b)})^\top \in \mathbb{R}^{n_b\times d}$ for $b=0, 1$ denote two batches from the same dataset. OT-tSNE modifies only t-SNE's high-dimensional similarities, while leaving the low-dimensional Cauchy similarities and the objective unchanged. To define these similarities jointly across the two batches, we construct an affinity matrix
$A\in\mathbb{R}^{(n_0+n_1)\times(n_0+n_1)}$ and set
$p_{ij} = A_{ij} / \sum_{kl}A_{kl}.$
The matrix $A$ has a four-block structure, with within-batch affinities on the diagonal and cross-batch affinities in the off-diagonal blocks. Fig.~\ref{fig:method} shows the schematic of OT-tSNE, and Algorithm~\ref{alg:ot-tsne} gives the pseudocode for its default configuration.

\textbf{Cross-batch affinities.}
The cross-batch affinity $P \in \mathbb{R}^{n_0\times n_1}$ is given by the entropic optimal transport (EOT) plan between $X_0$ and $X_1$ ~\citep{cuturi2013sinkhorn,gabriel2019computational}. Let \mbox{$C_{ij}=\lVert x_i^{(0)}-x_j^{(1)}\rVert_2^2$} denote the Euclidean cross-batch cost matrix. The EOT plan is defined as
\begin{equation}
\label{eq:eot}
P = \argmin_{\Gamma \geq 0}
\sum_{ij}C_{ij}\Gamma_{ij}
+ \epsilon \sum_{ij}\Gamma_{ij}\log\Gamma_{ij}
\quad \text{s.t.} \quad
\Gamma\mathbf 1=\frac{1}{n_0}\mathbf 1,\;
\Gamma^\top\mathbf 1=\frac{1}{n_1}\mathbf 1.
\end{equation} 
The resulting plan forms the off-diagonal blocks of the joint affinity matrix in Fig.~\ref{fig:method}. The marginal constraints ensure that the cross-batch coupling $P$ is globally balanced, so that all points are equally associated with the other batch. This is in contrast to local nearest-neighbor approaches that 
can lead to imbalanced associations with hubs and barely connected regions (Fig.~\ref{fig:fig1}d,e).

The hyperparameter $\epsilon$ controls the sparsity of the transport plan, with larger (smaller) $\epsilon$ yielding more uniform (concentrated) plans. We set $\epsilon=r\operatorname{median}_{ij} C_{ij}$, with $r=1$, following \citet{landa2024entropic}. We refer to the hyperparameter $r$ as relative regularization. 

\textbf{Within-batch affinities.} 
Our within-batch affinities, $S_0$, $S_1$, are inspired by EOT eigenmaps~\citep{landa2024entropic}, which leverage singular value decomposition of the cross-batch transport plan $P$
\begin{equation}
    P = U\Sigma V^\top,\qquad
    U=[u_1,\ldots,u_q],\quad
    V=[v_1,\ldots,v_q],\quad
    \Sigma=\operatorname{diag}(\sigma_1,\ldots,\sigma_q),
\end{equation}
where $q=\min(n_0,n_1)$, $u_i\in\mathbb{R}^{n_0}$,
$v_i\in\mathbb{R}^{n_1}$, and $\sigma_i$ are the left / right singular vectors, and singular values of $P$, respectively.
The EOT eigenmap coordinates are then given by the first columns of
\begin{equation}
    E =
    \begin{pmatrix}
        \sqrt{n_0}U\\
        \sqrt{n_1}V
    \end{pmatrix}.
\end{equation} 
We construct a symmetric joint affinity matrix $A$ whose eigenvectors are EOT eigenmap coordinates, while retaining $P$ as its cross-batch affinities. Specifically, we set
\begin{equation}
    A =\frac{1}
    {\sqrt{n_0n_1}}
    E\Sigma E^\top  =     \begin{pmatrix}
        S_0 & P\\
        P^\top & S_1
    \end{pmatrix} \text{ with }     S_0=\sqrt{{n_0}/{n_1}}U\Sigma U^\top,
    S_1=\sqrt{{n_1}/{n_0}}V\Sigma V^\top.
\end{equation}

We use this EOT eigenmap-based completion as our default. 

We consider four additional constructions of the within-batch affinities. \textbf{(i)} Setting the within-batch affinities to zero. \textbf{(ii)} Gram matrices $S_0 =PP^\top$ and $S_1=P^\top P$ relate points within one batch by how many cross-batch assignments they share. 
From the singular value decomposition, they satisfy
\begin{equation}
    PP^\top=U\Sigma^2U^\top,
    \qquad
    P^\top P=V\Sigma^2V^\top.
\end{equation}
Therefore, unlike the default completion above, this gives different powers of the singular values in the within- and cross-batch blocks. We refer to this method as OT-tSNE with Gram diagonals. \textbf{(iii)} Another optimal transport-based affinity matrix is the entropic optimal transport plan from each batch to itself. 
We refer to this variant as self OT. \textbf{(iv)} Finally, we represent within-batch affinities using a shared k-nearest-neighbor graph. This is closest to neighbor embedding methods like t-SNE and UMAP. We compare the alternative within-batch constructions empirically in Section~\ref{sec:results}.

\textbf{Batch-size-aware sparsification.}
The EOT plan, and consequently $A$, is dense, whereas the high-dimensional t-SNE affinities are sparse in practice. We therefore remove self-edges and select $k$ neighbors per observation, using $k=15$ as a practical default following UMAP and NCVis~\citep{mcinnes2018umap,artemenkov2020ncvis}. To avoid overrepresenting the larger batch, the neighbor budget is split according to the sizes of batches $b=\{0, 1\}$,
\begin{equation}
    k_b \approx k\frac{n_b}{n_0+n_1},
    \qquad k_0+k_1=k.
\end{equation}
Thus, for every observation, we keep its strongest $k_0$ affinities to batch~0 and strongest $k_1$ affinities to batch~1. We symmetrize the selected edges by retaining an edge whenever either endpoint selects the other. We then binarize the resulting graph and normalize its total affinity mass to one. The sparsification can introduce mild imbalance in the resulting affinity. 

\textbf{Initialization and optimization.}
By default, we initialize the 2D layout with the second and third columns of the EOT eigenmap matrix $E$. These coordinates incorporate cross-batch affinities into the initial layout. We compare four initialization choices (random, spectral, PCA, EOT), details on which can be found in Appendix~\ref{app:otsne algorithm}.

We then optimize the t-SNE objective using the sparse joint affinity matrix and this initial layout in \texttt{openTSNE}~\citep{polivcar2024opentsne}. Details on t-SNE parameters and other implementation details are in Appendix~\ref{app:otsne algorithm}.

\section{Experimental Setup}
\label{sec:setup}
\subsection{Datasets}
In addition to a synthetic version of MNIST~\citep{lecun1998gradient} with batch effect, described in Section~\ref{sec:synth_exp}, we benchmark OT-tSNE and competitors on seven real-world single-cell datasets with batch effect. We collected these from prior works on batch effect removal~\citep{tran2020benchmark, landa2024entropic} and added a subset of \citet{yao2021taxonomy}'s mouse brain atlas. The datasets vary in size from $\approx 4000$ to $\approx 30\, 000$ points. The batch effect ranges from replicated identical experimental conditions over varying biological signal (donor sex, treatment perturbation) to different single-cell measurement techniques. More details on the datasets can be found in Appendix~\ref{sec:data} and Tab.~\ref{tab:datasets}.

\subsection{Evaluation}
We evaluate each joint embedding along two complementary objectives: batch mixing and conservation of biological structure, following ~\citet{luecken2022benchmarking}. For each objective, we use two metrics, all scaled to $[0,1]$ such that higher values indicate better performance.

\textbf{Batch mixing.} We quantify batch mixing with the kNN batch-effect test~\citep[kBET]{buttner2019test} and the integration local inverse Simpson's index~\citep[iLISI]{korsunsky2019harmony}.

\textbf{Biological conservation.}
We measure preservation of biological structure using neighborhood purity and the silhouette score~\citep{rousseeuw1987silhouettes}.
We provide the precise metric definitions and implementation details in Appendix~\ref{app:metrics}.

\textbf{Ranking.}
Following established practices for benchmark aggregation~\citep{maierhein2018rankings,luecken2022benchmarking}, we aggregated metric values before ranking methods. Within each dataset, we min--max normalized each metric across methods. We ranked methods by \textbf{(i)} the mean of all four normalized metrics overall and \textbf{(ii)} by the mean of the two metrics in each category for biological conservation and batch mixing. We averaged the resulting ranks across the seven datasets. Additionally, we count how often a method obtains a top-3 rank per metric category and dataset. Appendix~\ref{app:ranking} provides further details on the ranking procedure and its interpretation.

\subsection{Compared methods} We compare OT-tSNE against prominent methods for batch integration and joint visualization, grouped according to how the final two-dimensional embedding is obtained. Tab.~\ref{tab:hyperparameters} summarizes the hyperparameters and configurations of the compared methods.

\textbf{Direct integration and visualization.}
We first consider methods dedicated to batch-corrected visualization. They directly construct a joint
two-dimensional representation and therefore provide the closest
comparison to OT-tSNE. These include MultiMAP~\citep{jain2021multimap}, BC t-SNE~\citep{aliverti2020projected}, RC t-SNE~\citep{heiter2023rctsne}, and Ref t-SNE ~\citep{policar2023embedding}.

\textbf{General-purpose integration methods.}
We additionally evaluate prominent general-purpose integration and
alignment methods: Scanorama~\citep{hie2019scanorama}, Seurat~\citep{stuart2019seuratcca}, scVI~\citep{lopez2018scvi}, LIGER~\citep{welch2019liger}, Harmony~\citep{korsunsky2019harmony}, and EOT eigenmaps ~\citep{landa2024entropic}. We evaluate these methods in two regimes. First, we use each method to construct a higher-dimensional integrated representation, as is common for downstream analyses. We choose 50 dimensions in our experiments before visualizing them with t-SNE. Second, we construct the integrated representation directly in 2D, a more restrictive setting that tests whether it can serve as the final visualization. We denote these two regimes by the method name followed by 50D or 2D, respectively, and refer to EOT eigenmaps simply as EOT throughout the experiments.

\textbf{No integration.}
Finally, we include joint PCA and joint t-SNE~\citep{maaten2008visualizing} as reference baselines, simply by concatenating the two batches and embedding them.

\section{Results}
\label{sec:results}
\subsection{Synthetic experiments}
\label{sec:synth_exp}
We first highlight the strength of optimal-transport-based cross-batch affinities. Nearest-neighbor affinities can become unreliable when faced with challenging batch effects. For instance, a well-known failure case is the ``kissing problem'', which is most immediate when the batch effect acts along signal directions~(Fig.~\ref{fig:fig1}c--f). In high-dimensional data, however, the structure of interest often lies on a much lower-dimensional manifold, such that batch effects are often orthogonal to the biological signal. This assumption underlies several batch correction techniques~\citep{haghverdi2018mnn, stuart2019seuratcca, hie2019scanorama}.
Here, we demonstrate that even in this setting, nearest-neighbor cross-batch affinities can struggle, while OT-tSNE is provably unaffected.

\textbf{Balanced OT is invariant to orthogonal batch effects.} 
Consider perturbing two datasets
by nuisance vectors, $\widetilde{x}^{(0)}_i=x^{(0)}_i+n_i^{(0)}$ and
$\widetilde{x}^{(1)}_j=x^{(1)}_j+n_i^{(1)}$. Assume that the nuisances are orthogonal to the data and to nuisances of the other dataset, i.e., all $n_i^{(0)}, n_j^{(1)}\in\mathbb{R}^d$ are orthogonal to all $x_i^{(0)}$ and $x_j^{(1)}$ and also each $n_i^{0}$ is orthogonal to each $n_j^{(1)}$.
Then, the squared Euclidean cost used in Eq.~\ref{eq:eot}  only changes by row- and column-wise offsets
\begin{equation}
    \widetilde C_{ij}
=
    C_{ij}
    + \|n_i^{(0)}\|_2^2
    + \|n_j^{(1)}\|_2^2
    + 2\left\langle x_i^{(0)} - x_j^{(1)},\, n_i^{(0)} - n_j^{(1)} \right\rangle 
    - 2\left\langle n_i^{(0)}, n_j^{(1)} \right\rangle
    =:
    C_{ij}+\nu_i^{(0)}+\nu_j^{(1)},
    \label{eq:orthogonal_cost}
\end{equation}
because all other terms are zero due to orthogonality. For any feasible coupling $\Gamma$ in Eq.~\ref{eq:eot}, the perturbed and unperturbed cost functions differ by a constant, 
\begin{equation}
    \sum_{ij}\widetilde C_{ij}\Gamma_{ij}
    =
    \sum_{ij}C_{ij}\Gamma_{ij}
    + \frac{1}{n_0}\sum_i \nu_i^{(0)}
    + \frac{1}{n_1}\sum_j \nu_j^{(1)},
    \label{eq:ot_row_column_invariance}
\end{equation}
since the additional terms are fixed by the marginals. Hence, for fixed $\epsilon$, the optimal EOT plan $P$ is unchanged under this perturbation model. More general orthogonal nuisance models have been theoretically analyzed by \citet{kuang2017preconditioning} and \citet{landa2024entropic}. 

In contrast, nearest-neighbor affinities are sensitive to the nuisance magnitudes in Eq.~\ref{eq:orthogonal_cost}, which can reorder the pairwise distances and hence change the local neighborhoods.

\begin{figure}
    \centering
    \includegraphics[width=\linewidth]{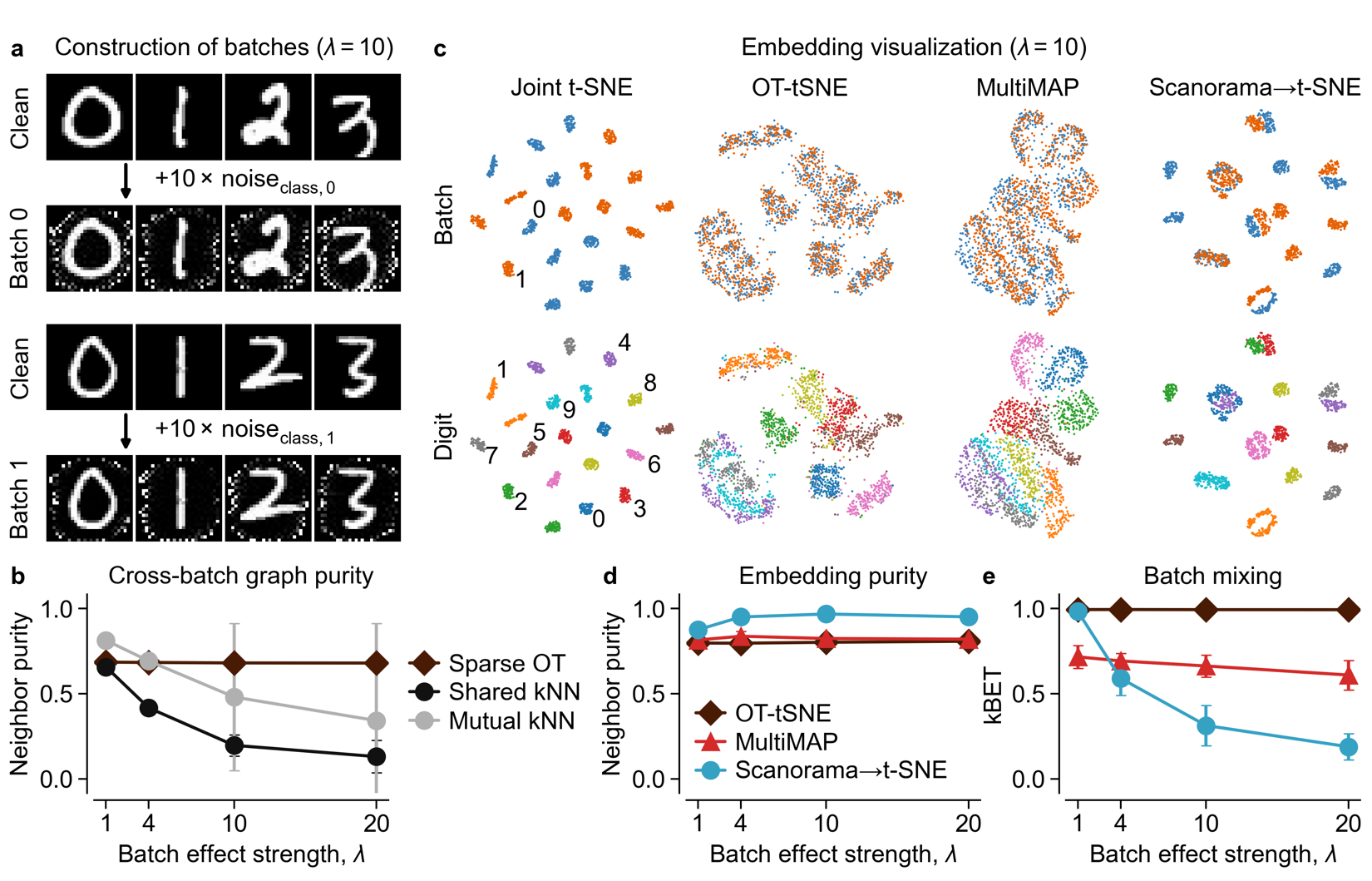}
    \caption{\textbf{a.} Construction of class-dependent orthogonal batch effects. \textbf{b.} Cross-batch graph purity for OT, shared kNN, and mutual kNN affinities. \textbf{c.} Example embeddings at $\lambda=10$, colored by batch (top) and digit class (bottom). \textbf{d--e.} Embedding purity and batch mixing for the select integrated visualization methods across corruption strengths.}
    \label{fig:synthetic_experiment}
\end{figure}

\textbf{MNIST with orthogonal class-dependent batch effects.}
We apply the above noise setting to two disjoint, class-balanced batches sampled from MNIST. From a joint PCA of the clean images, we use two disjoint sets of 30 trailing components to define mutually orthogonal noise subspaces (Appendix~\ref{app:synthetic_construction}). Specifically, each digit class receives its own perturbation direction and magnitude,
\begin{equation}
    \widetilde{x}_i^{(0)}
    =
    x_i + \lambda n_{c(i)}^{(0)},
    \qquad
    \widetilde{x}_j^{(1)}
    =
    x_j + \lambda n_{c(j)}^{(1)},
    \label{eq:mnist_corruption}
\end{equation}
where $c(i)$ denotes the digit class of sample $i$ and $\lambda$ controls the batch-effect strength (Fig.~\ref{fig:synthetic_experiment}a). Thus, different regions of the data experience batch effects of different directions and magnitudes.
This is different from the usual orthogonality notion, where one batch is rigidly shifted relative to the other by some shared batch vector orthogonal to the signal subspace. Nevertheless, in our setup the batch effect is still confined to dimensions orthogonal to the main signal.  Since we retain the original pixels as the clean signal and choose the perturbation from the trailing PCs, the orthogonal decomposition in Eq.~\ref{eq:orthogonal_cost} is only approximate. Projecting the images onto the leading PCA subspace before adding noise makes this orthogonality exact and yields nearly identical results (Fig.~\ref{app:fig:reg_orthog}c,g).

\textbf{Regularization.}
At each corruption strength $\lambda$, we recomputed our default regularization $\epsilon_\lambda=\operatorname{median}_{ij} C_\lambda(i,j)$, which increased with $\lambda$ (Fig.~\ref{app:fig:reg_orthog}a). The invariance result assumes fixed $\epsilon$. In practice, cross-batch digit-label purity remained similar under fixed and adaptive regularization (Appendix~\ref{app:synhtetic_reg}, Fig.~\ref{app:fig:reg_orthog}c).

\textbf{Cross-batch affinity stability.} We compared sparsified EOT affinities with shared and mutual cross-batch kNN affinities using the same neighborhood size $k=15$. For EOT affinities, we retained the top-$15$ cross-batch affinities per point, symmetrized the resulting graph and binarized it. We evaluated the affinities by the neighbor purity, defined as the fraction of retained neighbors with the same digit label, averaged over points with at least one retained edge (Fig.~\ref{fig:synthetic_experiment}b). As the strength of the batch effect increased, both shared and mutual kNN affinities progressively deteriorated. For mutual kNN, this deterioration persisted despite conservative matching, as the fraction of points retaining a reciprocal cross-batch edge fell from $98\%$ to $11\%$ between $\lambda=0$ and $\lambda=20$. In contrast, the EOT affinities remained stable over the same corruption range. 

\textbf{Two-dimensional embeddings.} We finally ask whether this difference in cross-batch affinities affects the resulting two-dimensional embeddings. We compare OT-tSNE with MultiMAP~\citep{jain2021multimap}, whose integration graph uses shared cross-batch neighborhood information, and Scanorama~\citep{hie2019scanorama} followed by t-SNE, which relies on mutual nearest-neighbor correspondences. OT-tSNE maintains nearly unchanged batch mixing and digit purity across the corruption range (Fig.~\ref{fig:synthetic_experiment}d--e). Scanorama progressively deteriorates in batch mixing, while MultiMAP is more robust but also deteriorates as the batch effect increases. Digit purity remains high for all three methods, indicating that the dominant effect is a loss of cross-batch alignment. The visualization of embeddings at $\lambda=10$ in Fig.~\ref{fig:synthetic_experiment}c qualitatively reflects these trends.

Together, these experiments show that under heterogeneous orthogonal batch
effects, balanced transport yields more stable cross-batch affinities than local neighbor matching, and that this difference persists in the resulting two-dimensional visualizations.

\begin{figure}[tb]
    \centering
    \includegraphics[width=\linewidth]{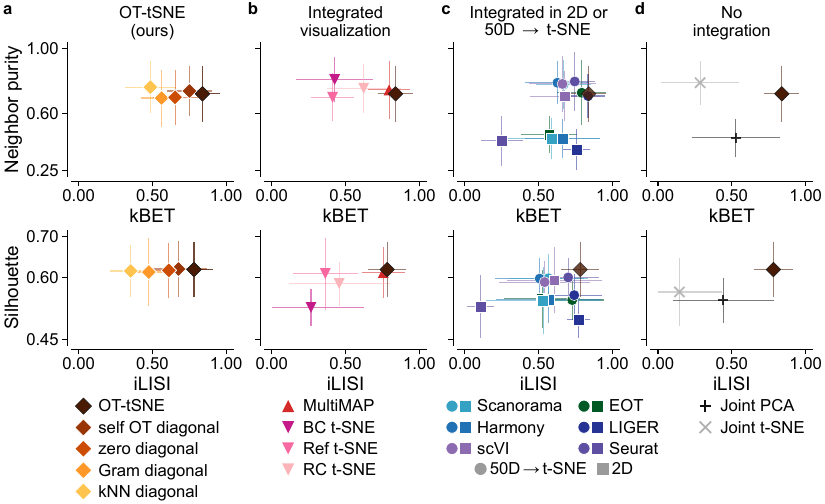}
    \caption{Pooled batch-mixing scores (x-axis; kBET or iLISI) versus biological-conservation scores (y-axis; purity or silhouette) across datasets for \textbf{a.} OT-tSNE variants, \textbf{b.} integrated visualization methods, \textbf{c.} general-purpose integration methods either with 50D latent spaces visualized with t-SNE (dots) or with 2D latent spaces directly used for visualization (squares), and \textbf{d.} baselines without integration. Reported means and standard deviations over datasets. OT-tSNE is repeated in all panels for better comparability.}
    \label{fig:results_pooled_benchmarking}
\end{figure}

\subsection{Benchmarking on real datasets}
\label{sec:results_benchmark}

To test whether the results in the controlled setting extend to real data, we extensively benchmarked OT-tSNE on seven single-cell datasets with diverse batch effects and varying degrees of compositional imbalance (Figs.~\ref{app:fig:batch_composition} and ~\ref{app:fig:batch_composition_2}). 
For instance, Mouse Atlas showed the second-largest compositional imbalance in our benchmark, with its B cells accounting for roughly 29\% of one batch and 8\% of the other (Fig.~~\ref{app:fig:batch_composition}c). Remarkably, despite this disparity, OT-tSNE had the best trade-off in mixing and biological preservation in Mouse Atlas compared to the other methods (Figs.~\ref{app:fig:benchmark_dataset_kbet_purity} and ~\ref{app:fig:benchmark_dataset_ilisi_silhouette}).

\begin{wraptable}{r}{0.3\textwidth}
 \vspace*{-0.35cm}
  \centering
  \caption{Aggregated ranks across all four metrics and all datasets, lower is better.}
  \label{tab:ranks}
\begin{tabular}{lr}
\toprule
Method & Rank \\
\midrule
OT-tSNE             &  2.0 \\
OT-tSNE (self OT)   &  4.0 \\
MultiMAP            &  4.4 \\
Seurat 50D          &  5.4 \\
LIGER 50D           &  7.7 \\
OT-tSNE (zero)      &  8.0 \\
Scanorama 50D       &  8.6 \\
EOT 50D             &  9.0 \\
scVI 50D            &  9.4 \\
Harmony 50D         &  9.6 \\
RC t-SNE            & 10.4 \\
scVI 2D             & 11.3 \\
OT-tSNE (Gram)      & 11.4 \\
OT-tSNE (kNN)       & 12.7 \\
Ref t-SNE           & 16.1 \\
Harmony 2D          & 16.3 \\
LIGER 2D            & 16.7 \\
BC t-SNE            & 16.9 \\
Joint t-SNE         & 17.4 \\
Scanorama 2D        & 17.9 \\
EOT 2D              & 18.9 \\
Joint PCA           & 19.4 \\
Seurat 2D           & 22.4 \\
\bottomrule
\end{tabular}
\vspace*{-1.1cm} 
\end{wraptable} 

\begin{figure}
    \centering
    \includegraphics[width=\linewidth]{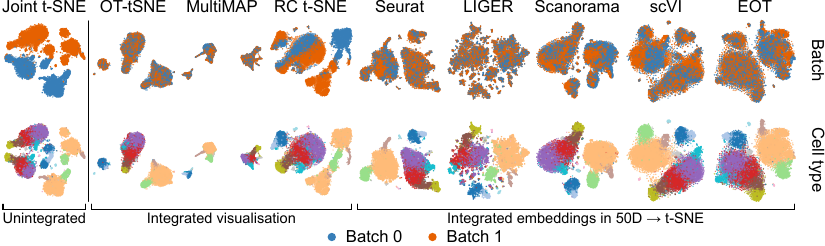}
    \caption{Lupus embeddings across selected methods. Top: colored by batch. Bottom: the same embeddings colored by annotated cell type. For visualizations of the other datasets, see Fig.~\ref{app:fig:embeddings_other_datasets}.}
    \label{fig:embeddings_lupus}
    \vspace*{-0.25cm}
\end{figure}

Across the seven datasets, OT-tSNE overall achieved a strong balance between batch mixing and biological conservation. Its pooled scores were among the best methods for all metrics, with particularly strong batch mixing scores, kBET and iLISI (Fig.~\ref{fig:results_pooled_benchmarking}). This performance was also consistent across datasets and metrics: OT-tSNE appears among the top three ranked methods more often than any competing method (Fig.~\ref{fig:fig1}g). The aggregate ranking over all four metrics likewise placed OT-tSNE first (Tab.~\ref{tab:ranks}). 

The construction of the within-batch affinities mostly affected batch mixing. The five OT-tSNE variants had similar pooled purity and silhouette scores, whereas the default spectral construction attains the highest kBET and iLISI scores (Fig.~\ref{fig:results_pooled_benchmarking}a). We therefore use this construction as the default throughout the paper.

MultiMAP was the strongest competitor (Fig.~\ref{fig:results_pooled_benchmarking}b). It was slightly better in biological conservation, while OT-tSNE mixed batches better (Fig.~\ref{app:fig:ranks}). The other integrated visualization tools, BC~t-SNE, RC~t-SNE, and Ref t-SNE, integrated batches much worse.

General-purpose integration methods followed by t-SNE were competitive but less consistent across datasets (Fig.~\ref{fig:results_pooled_benchmarking}c). Some 50-dimensional integrations matched or exceeded OT-tSNE on individual biological-conservation metrics, but their performance varied substantially across metrics and datasets (Figs.~\ref{app:fig:benchmark_dataset_kbet_purity} and ~\ref{app:fig:benchmark_dataset_ilisi_silhouette}). Adapting the methods to output integrated embeddings in 2D generally reduced biological conservation relative to integrating in 50 dimensions followed by t-SNE, with scVI-2D as a notable exception.

Without batch integration, joint t-SNE largely preserved biological structure but left the batches separated. In contrast, joint PCA mixed the batches at the expense of biological conservation (Fig.~\ref{fig:results_pooled_benchmarking}d).

Embeddings of the selected methods on the Lupus dataset illustrate the corresponding trade-off between batch mixing and cell-type structure (Fig.~\ref{fig:embeddings_lupus}).

The main exception was the Hematopoietic dataset, where some cell types occurred in only one batch (Fig.~\ref{app:fig:batch_composition}b). Here, OT-tSNE retained comparatively strong batch mixing but lost more biological structure than on the other datasets (Figs.~\ref{app:fig:benchmark_dataset_kbet_purity} and
\ref{app:fig:benchmark_dataset_ilisi_silhouette}). However, this dataset was generally more challenging to integrate for all the methods.

\textbf{Sensitivity to hyperparameters and initialization.}
We evaluated 25 combinations of neighborhood size $k$ and relative regularization $r$ on the seven real datasets (Appendix~\ref{app:hyperparameter_sensitivity}). On average, batch mixing and biological conservation changed little near the default $(k,r)=(15,1)$, with larger trade-offs at the extremes of the tested range. At fixed $k$, lower regularization tended to favor batch mixing, while at fixed $r$, larger neighborhoods tended to favor biological conservation (Figs.~\ref{app:fig:hyperparameter_pooled} and ~\ref{app:fig:hyperparameter_lines}). 
Across all five OT-tSNE variants, switching from random initialization to PCA, spectral, or EOT initialization changed each metric by less than 0.01 on average across datasets (Figs.~\ref{app:fig:init_pooled} and ~\ref{app:fig:init_datasets}). More details are in Appendix~\ref{app:initialization_sensitivity}. On Lupus, random initialization produced different cluster arrangements across seeds, whereas each of the deterministic initializations (spectral, EOT, and PCA) reproduced the same visualization when jitter was disabled (Fig.~\ref{app:fig:init_seeds_lupus_nojitter}).

\section{Limitations and Future Work}
\label{sec:limitations}
OT-tSNE uses balanced optimal transport with uniform marginals. When a cell type is more abundant in one batch than in the other, some of its mass must be transported to other cell types. This can be problematic when batches contain strongly different cell-type proportions or non-overlapping cell types. Therefore, we assume that this constraint may have contributed to the weaker cell-type preservation we observed on the Hematopoietic dataset, where some cell types occur in only one batch (see Fig.~\ref{app:fig:batch_composition}b for batch composition, Fig~\ref{app:fig:benchmark_dataset_kbet_purity} and ~\ref{app:fig:benchmark_dataset_ilisi_silhouette}). Unbalanced optimal transport~\citep{chizat2018scaling} provides a possible extension by relaxing the marginal constraints and thereby avoiding forced correspondences between populations that are absent from the other batch. Density-reweighted EOT~\citep{li2026density} offers another possible extension by reducing the influence of sampling-density differences on cross-batch correspondences.

OT-tSNE is currently formulated for two batches. An easy multi-batch extension would compute pairwise transport plans between batches as off-diagonal blocks of a larger joint affinity matrix.

Runtime is another limitation. In our CPU-only benchmark, default OT-tSNE was comparable to several competing methods on smaller datasets, but the affinity matrix creation increasingly dominated as dataset size grew, whereas the other OT-tSNE variants were faster on larger datasets (Fig.~\ref{app:fig:runtime}).  The SVD and construction of within-batch affinities were the main computational bottlenecks in the default OT-tSNE, which can be easily sped up with approximate versions. Appendix~\ref{app:comp_complexity} analyzes these costs and summarizes the computational complexity of the benchmarked methods. We therefore tested randomized SVD truncated to 200 components, which substantially reduced runtime and made OT-tSNE comparable to several competing methods also on larger datasets (Fig.~\ref{app:fig:runtime}), while preserving its first-place aggregate ranking. Beyond that, EOT's Sinkhorn iterations~\citep{cuturi2013sinkhorn} can be accelerated on GPU. Landmark approximations~\citep{altschuler2019massively} could further reduce the size of the transport problem.

Finally, unlike general-purpose integration methods, OT-tSNE produces a two-dimensional visualization rather than a reusable integrated representation for downstream analysis. A promising extension is to use OT-derived cross-batch correspondences as supervision for high-dimensional representation learning, for example through a contrastive objective.

\section{Conclusion}
We presented OT-tSNE for visualization of datasets with batch effects. Rather than using local cross-batch nearest-neighbor relations, OT-tSNE uses entropic optimal transport to construct high-dimensional cross-batch affinities directly visualized by t-SNE. Our controlled experiments showed that these affinities were robust under heterogeneous orthogonal batch effects, where nearest-neighbor affinities deteriorate. In a benchmark on seven real single-cell datasets, OT-tSNE was fairly robust even when datasets with shared classes had unbalanced distributions. In this benchmark, OT-tSNE consistently achieved strong batch mixing while preserving biological structure well. 

\section{Acknowledgments}

This work was supported by the Gemeinnützige Hertie-Stiftung. We thank the International Max Planck Research School for Intelligent Systems (IMPRS-IS) for supporting  Azhar Akhmetova.

We thank Florian Beier and Stephan Eckstein for productive discussions on unbalanced optimal transport.

\subsection*{AI use statement}

In this work, we used generative AI tools for generating synthetic datasets (Appendix~\ref{app:synthetic_construction}), designing or providing feedback on research methodology and experiments, implementing methods, cleaning and reformatting datasets (esp. joint preprocessing in Appendix~\ref{app:preprocessing}). We did not use generative AI tools to develop theoretical models or
conceptual frameworks, formulate mathematical claims, propose or refine hypotheses, interpret results, or provide critical ingredients for proving mathematical claims. Assistance with writing proofs, qualitative or thematic data analysis, and translation were not applicable to this work.

Additionally, we used generative AI tools for creating and editing software code, drafting and editing parts of the manuscript, and formatting references. We manually edited all text afterwards. Similarly, all AI-assisted code and analyses were reviewed by the authors. We take responsibility for the final content of this work, including text, claims, code, analyses, and artifacts produced with the aid of generative AI.

\subsection*{Reproducibility statement}
We ensure reproducibility by clearly describing our method (Section~\ref{sec:method}) and experiment setup, including datasets, metrics and competitors (Section~\ref{sec:setup}) in the main paper. Additionally, we give further details on the single-cell datasets and preprocessing in Appendix~\ref{sec:data} and describe the synthetic dataset construction in Appendix~\ref{app:synthetic_construction}. We dedicate Appendix~\ref{app:implementation_details} to implementation details of the benchmarked methods. Overall, we stuck to default hyperparameters as far as possible. We give full details on the evaluation metrics in Appendix~\ref{app:metrics}. Finally, we include our code as supplementary and will make it open source upon acceptance.

\bibliography{references}
\bibliographystyle{plainnat}
\clearpage
\appendix
\renewcommand{\thefigure}{S\arabic{figure}}
\renewcommand{\theHfigure}{S\arabic{figure}}
\setcounter{figure}{0}  
\renewcommand{\thetable}{S\arabic{table}}
\renewcommand{\theHtable}{S\arabic{table}}
\setcounter{table}{0} 
\section*{Appendix}

\section{Single-cell dataset descriptions}
\label{sec:data}
All but the Human brain datasets contain single-cell RNA-sequencing data. 

\paragraph{Human brain} contains single-nucleus RNA-seq data for $n=9215$ cells from the same human postmortem insular-cortex sample, profiled using 10x Chromium v3 snRNA-seq~\citep{ament2024single}. The two batches correspond to replicates 1 and 4 and the batch effect is only due to technical artifacts of measuring both batches independently with the same protocol. It is therefore quite small. We downloaded the data from \url{https://zenodo.org/records/17167270}.
\paragraph{nCov} contains $n=4117$ peripheral blood mononuclear cells (PBMCs) from a longitudinal study of a COVID-19 patient 5 of~\citet{zhu2020single}. The batches were sampled at days 1 and 13 after hospitalization and so that the batch effect is mainly due to  the disease progression.
\paragraph{Lupus} contains $n=13\,000$ PBMCs from lupus patients~\citep{kang2018multiplexed}. The batches correspond to the control and treatment cohorts.
\paragraph{Mouse Cortex} contains the $n=26\,588$ layer 2/3 intratelencephalic-projecting glutamatergic neurons of the entorhinal mouse cortex in~\citet{yao2021taxonomy}, i.e., subclasses L2 IT ENTl, L2 IT ENTm, L3 IT ENT, L2/3 IT ENTl. The batch effect corresponds to the sex of the mouse.
\paragraph{PBMC} contains $n=15\,476$ peripheral blood mononuclear cells of healthy donors recorded either with the 10x Chromium 3' or 5' chemistry from~\citet{zheng2017massively}. The batch effect is thus due to measurement technology.
\paragraph{Mouse atlas} contains $n= 6954$ cells from the two whole mouse atlases of~\citet{han2018mapping} and~\citet{tabula2018single}. Only 11 shared cell types are used following~\citet{tran2020benchmark}. This dataset contains a strong batch effect since \citet{han2018mapping} acquired the data via the Microwell-seq technology, while the \citet{tabula2018single} used SMART-seq2.
\paragraph{Hematopoietic} contains $n=4649$ differentiating hematopoietic stem and progenitor cells from mouse bone marrow. Again, the two batches come from different studies. The first batch is from~\citet{paul2015transcriptional} who used the MARS-seq protocol. The second is from~\citet{nestorowa2016single} who used  SMART-seq2. We used the same subsets as~\citet{tran2020benchmark}. In addition to the different measurement technologies, the two batches contain different cell types and even the shared ones have very different proportions (Figs.~\ref{app:fig:batch_composition} and ~\ref{app:fig:batch_composition_2}), making this a challenging dataset with strong batch effects.  

The last three datasets (PBMC, mouse atlas, and hematopoietic) were used in~\citet{tran2020benchmark} and we downloaded them from the author's Docker image \url{https://hub.docker.com/r/jinmiaochenlab/batch-effect-removal-benchmarking}.

\begin{table}[b]
\caption{Single-cell datasets overview}
\label{tab:datasets}
\begin{center}
\begin{tabular}{lllrrr}
\toprule
Name & Description & Main batch effect & $n_0$ & $n_1$ & classes\\
\midrule 
Human brain & human insular cortex & technical replicates &$3145$ & $6070$ & $9$ \\
nCov         &PBMCs of COVID patient & days in hospital &  $2877$ & $1240$ & $13$\\
Lupus & PBMCs of Lupus patients & treatment vs control & $7451$& $6548$ & $13$ \\
Mouse cortex &mouse entorhinal cortex  & donor sex & $11352$&$15236$ & $22$\\
PBMC & PBMCs & sc-technology & $8098$ & $7378$ & $9$ \\
Mouse atlas & mouse atlas & sc-technology & $4239$& $2715$ & $11$\\
Hematopoeitic &mouse hematopoiesis&sc-technology&$2729$& $1920$ & $7$\\
\bottomrule
\end{tabular}
\end{center}
\end{table}

\subsection{Preprocessing}
\label{app:preprocessing}
We used \texttt{scanpy} to apply a unified preprocessing pipeline to all datasets. Our aim was to standardize the input across methods as much as possible and thereby disentangle the effects of preprocessing from those of the integration methods themselves. First, we concatenated the batches and removed cells with fewer than 10 detected genes and genes detected in fewer than three cells. We selected $1\,000$ highly variable genes using the \texttt{seurat\_v3} flavor. 
Feature selection was performed on raw counts.

We retained the raw counts for the selected genes as input to methods that require count data: scVI~\citep{lopez2018scvi}, Seurat~\citep{stuart2019seuratcca}, LIGER~\citep{welch2019liger}, and Ref~tSNE~\citep{policar2023embedding}. For these methods, internal feature selection was disabled so that all methods used the same 1,000-gene panel, while other method-specific preprocessing of the input was retained.

For the remaining methods, we normalized each cell to a total count of $10\,000$ and applied the natural-log transformation $\log(1+x)$. Scanorama~\citep{hie2019scanorama}, MultiMAP~\citep{jain2021multimap}, and BC tSNE~\citep{aliverti2020projected} operated directly on this common log-normalized $1\,000$-gene representation, as these methods apply their own preprocessing in feature space. Finally, we fitted a single 50-dimensional PCA jointly to the concatenated batches to obtain a common PCA representation. OT-tSNE, EOT eigenmaps~\citep{landa2024entropic}, joint t-SNE, Harmony~\citep{korsunsky2019harmony}, and RC tSNE~\citep{heiter2023rctsne} operated on the corresponding batches in this 50-dimensional representation. Overall, all methods used the same cells and selected genes while receiving the representation appropriate for their input requirements.

\subsection{Batch compositions}
\label{app:sec:batch compositions}

\begin{figure}[tb]
    \centering
    \includegraphics[width=\linewidth]{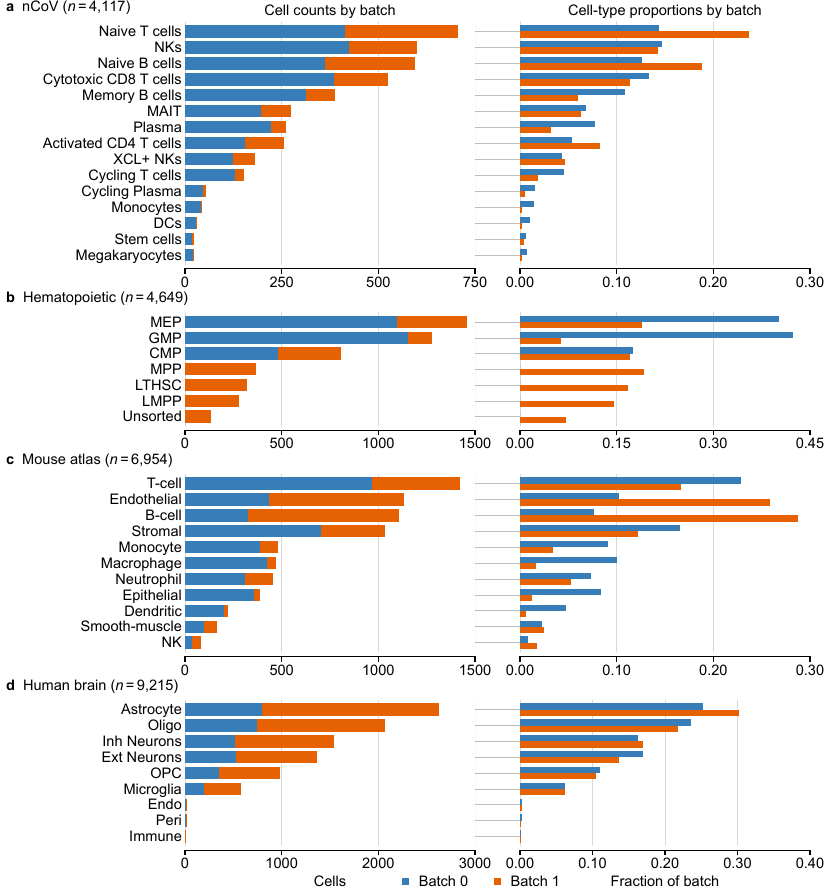}
    \caption{Batch compositions of the subset of single-cell datasets: nCoV, Hematopoietic, Mouse atlas, Human brain. Cell counts per cell type (left) and corresponding within-batch cell-type proportions (right).}
    \label{app:fig:batch_composition}
\end{figure}

\begin{figure}[tb]
    \centering
    \includegraphics[width=\linewidth]{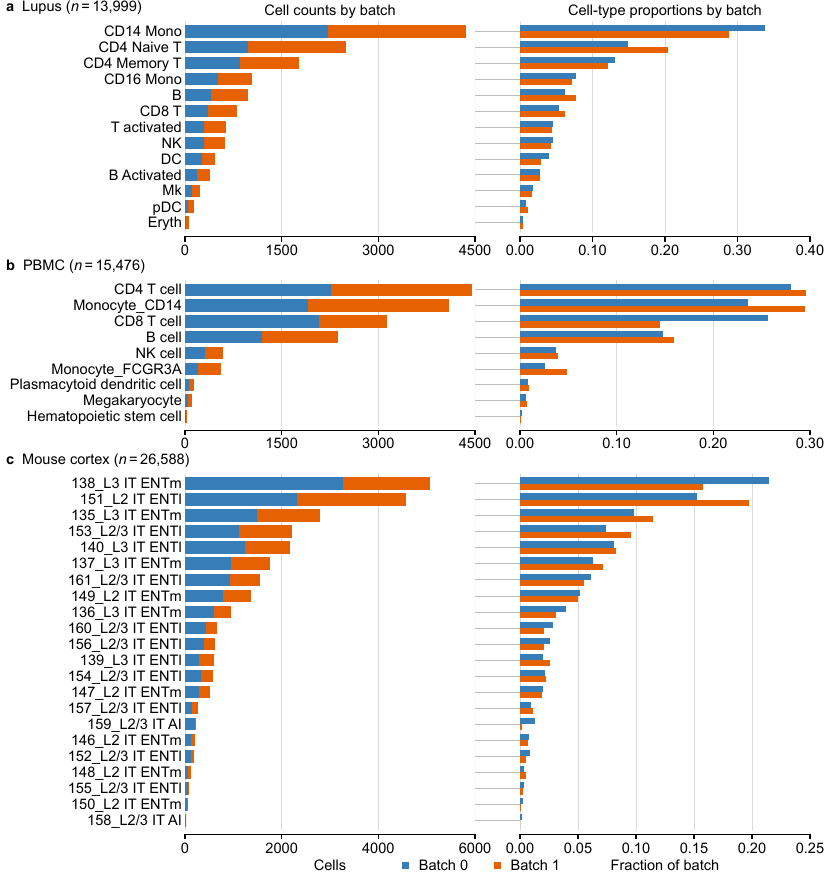}
    \caption{Batch compositions of the subset of single-cell datasets: Lupus, PBMC, Mouse cortex. Cell counts per cell type (left) and corresponding within-batch cell-type proportions (right).}
    \label{app:fig:batch_composition_2}
\end{figure}

The datasets vary in their cell-type composition across batches (Fig.~\ref{app:fig:batch_composition} and ~\ref{app:fig:batch_composition_2}). All datasets except hematopoietic contain shared cell types across batches, whereas hematopoietic additionally contains batch-specific cell types. Cell-type proportions are relatively balanced across batches in some datasets (Lupus, Human Brain, Mouse cortex, PBMC), but are rather imbalanced in the other datasets (nCov, Hematopoietic, Mouse atlas). Consequently, batch effects can be entangled with biological composition, making integration particularly challenging for datasets with imbalanced or non-overlapping cell types.


\clearpage
\section{Synthetic experiment}

We used the default configurations for MultiMAP and Scanorama. For the reported OT-tSNE embeddings, we used random initialization and kept all other settings at their default values.

\subsection{Synthetic orthogonal batch effect construction}
\label{app:synthetic_construction}
We sample two disjoint, class-balanced MNIST batches with 100 images per digit and fit PCA jointly to both batches. We take the last 60 numerically resolved PCs (explained variance \(>10^{-10}\) times that of the first PC) and split them into two disjoint 30-dimensional nuisance subspaces, one per batch.

For each class \(c\) and batch \(b\), we sample a unit nuisance direction
\begin{equation}
q_c^{(b)}
= \frac{\sum_{k=1}^{30} z_{ck}^{(b)}u_k^{(b)}}
{\left\|\sum_{k=1}^{30} z_{ck}^{(b)}u_k^{(b)}\right\|_2},
\qquad
z_{ck}^{(b)}\sim\mathcal N(0,1),
\end{equation}
where \(u_k^{(b)}\in\mathbb R^{784}\) are the corresponding PCA basis vectors. The class scales \(s_c^{(b)}\) are independent permutations, for each batch, of ten values equally spaced between \(0.35\) and \(2.60\). We define
\begin{equation}
n_c^{(b)}
=
s_c^{(b)}q_c^{(b)}
-\frac{1}{10}\sum_{r=0}^{9}s_r^{(b)}q_r^{(b)},
\end{equation}
so that the average nuisance within each batch is zero. The corrupted samples are
\(\widetilde x_i^{(b)}=x_i^{(b)}+\lambda n_{c(i)}^{(b)}\), with \(\lambda\in\{1,4,10,20\}\).

\subsection{Fixed and adaptive regularization in the MNIST experiment}
\label{app:synhtetic_reg}
\begin{figure}[tb]
    \centering
    \vspace*{-5mm}
    \includegraphics[width=\linewidth]{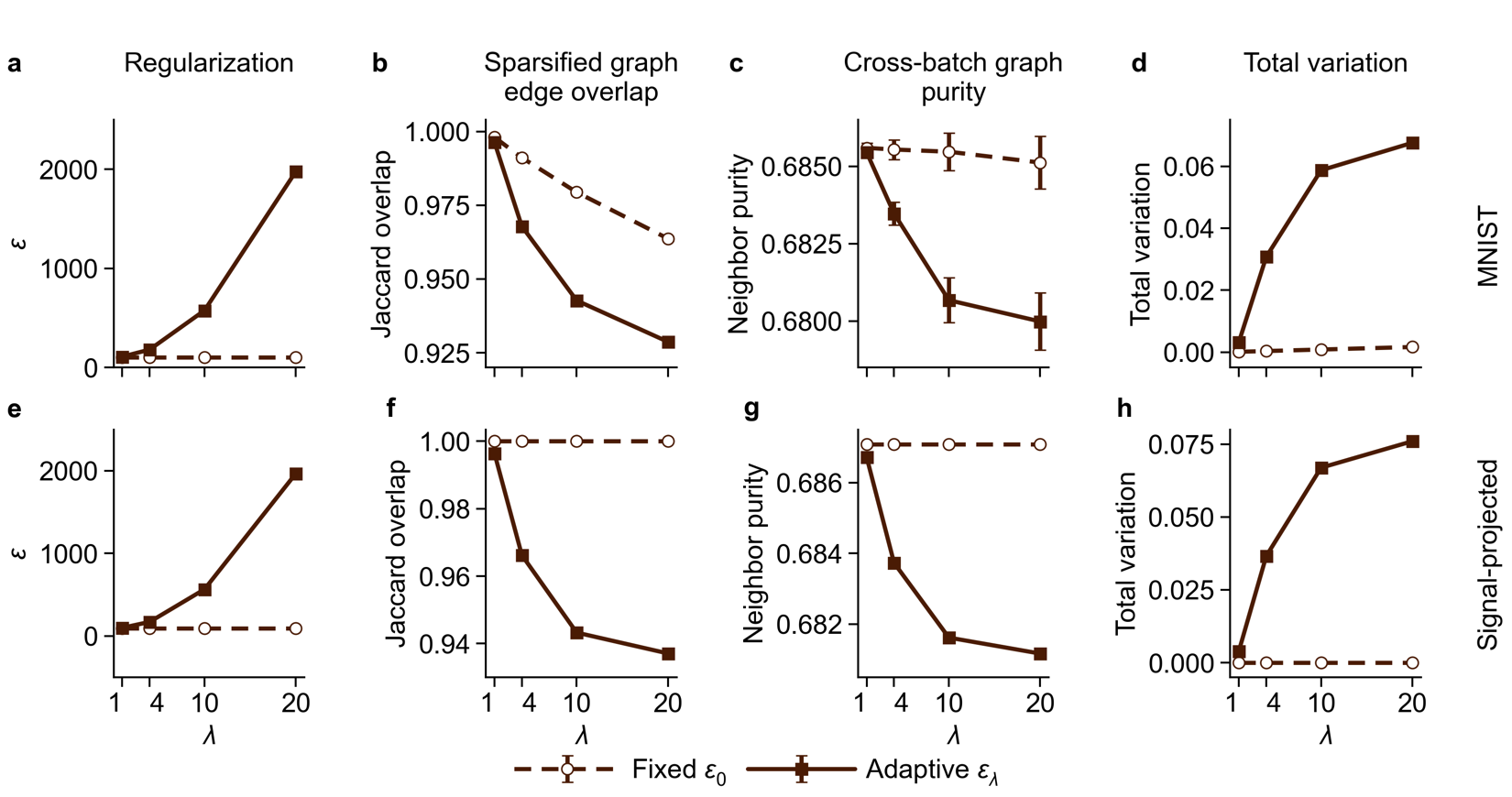}
\caption{Effect of fixed and adaptive entropic regularization on original MNIST (top) and the signal-projected control (bottom). Columns show regularization strength (a,e), graph edge overlap (b,f), digit-label purity (c,g), and full-plan total variation (d,h). Edge overlap and total variation are measured relative to zero corruption. Points and error bars show means and standard deviations over three seeds.}
    \label{app:fig:reg_orthog}
\end{figure}

In Fig.~\ref{fig:synthetic_experiment}, we used our default entropic regularization, $\epsilon_\lambda=\operatorname{median}_{ij} C_\lambda(i,j)$, where $C_\lambda$ is the cross-batch squared Euclidean cost matrix at corruption strength $\lambda$. This regularization increased with corruption strength, whereas the invariance result in Section~\ref{sec:synth_exp} assumes fixed regularization and fixed marginals. We therefore repeated the experiment with $\epsilon_0=\operatorname{median}_{ij} C_0(i,j)$ held fixed across corruption strengths (Fig.~\ref{app:fig:reg_orthog}a,e).

We compared fixed and adaptive regularization on the original MNIST data used in Fig.~\ref{fig:synthetic_experiment} and on a signal-projected control. For the control, we projected the clean images onto the leading principal components explaining at least $90\%$ of their pooled variance before adding the class-dependent corruption. This makes the corruption exactly orthogonal to the retained signal. On original MNIST, the clean images also contain variation outside this subspace, so orthogonality holds only approximately. 

To assess whether these changes affected the selected neighbors, we constructed a cross-batch graph by retaining the 15 largest entries in each row and each column of the plan. A pair forms an edge if either observation selects the other. We measured edge stability by the Jaccard overlap with the uncorrupted graph: the number of shared edges divided by the number of edges in their union. We also measured digit-label purity by computing the fraction of each observation's graph neighbors with the same digit label, averaging first within each batch and then equally across batches. We also measured changes in the dense transport plan by its total variation from the uncorrupted plan, i.e., the fraction of transport mass reassigned. 

At fixed $\epsilon_0$, the projected control behaved as predicted by Section~\ref{sec:synth_exp}. The full plan remained unchanged up to numerical precision, so the top-15 selections also remained unchanged and edge overlap was one at every corruption strength (Fig.~\ref{app:fig:reg_orthog}f,h). On original MNIST, we observed small changes in both the plan and the selected edges (Fig.~\ref{app:fig:reg_orthog}b,d).

With adaptive regularization, total variation at the strongest corruption ($\lambda=20$) reached 0.068 on original MNIST and 0.076 on the projected control (Fig.~\ref{app:fig:reg_orthog}d,h). Even at this strength, the respective graphs had Jaccard overlaps of 0.929 and 0.937 with their uncorrupted references (Fig.~\ref{app:fig:reg_orthog}b,f). Digit-label purity decreased by only about half a percentage point in both settings from $\lambda=1$ to $\lambda=20$ (Fig.~\ref{app:fig:reg_orthog}c,g).

\section{Implementation details}
\label{app:implementation_details}
\subsection{OT-tSNE algorithm}
\label{app:otsne algorithm}
\vspace*{-5mm}
\begin{algorithm}
\caption{OT-tSNE with spectral within-batch affinities and EOT initialization}
\label{alg:ot-tsne}
\begin{algorithmic}[1]
\Require Batches $X^{(b)} \in \mathbb{R}^{n_b \times d}$ for $b \in \{0,1\}$,
  relative regularization $r$ (default $1$), neighbor budget $k$ (default $15$),
  seed $s$
\Ensure Joint embedding $Y \in \mathbb{R}^{(n_0+n_1)\times 2}$

\Statex \textit{Cross-batch transport}
\State $C_{ij} \gets \lVert x_i^{(0)} - x_j^{(1)} \rVert_2^2$,
  \quad $\epsilon \gets r\,\operatorname{median}_{ij} C_{ij}$
\State $P \gets \Call{Sinkhorn}{C,\epsilon,
  \mathbf{1}_{n_0}/n_0,\mathbf{1}_{n_1}/n_1}$

\Statex \textit{Joint affinities}
\State Compute the SVD $P = U\Sigma V^\top$
\State $E \gets \begin{pmatrix}\sqrt{n_0}\,U \\ \sqrt{n_1}\,V\end{pmatrix}$,
  \quad $A \gets E\Sigma E^\top/\sqrt{n_0n_1}$
  \Comment{cross-batch blocks are $P$ and $P^\top$}
\State Set negative entries of $A$ to zero

\Statex \textit{Batch-size-aware sparsification}
\State Split $k$ into integer budgets $k_0+k_1=k$ with
  $k_b \approx k\,n_b/(n_0+n_1)$
\For{each observation $i$ and each batch $b \in \{0,1\}$}
  \State Select the $k_b$ observations $j$ in batch $b$ with the
    largest $A_{ij}$, excluding $i$ itself
\EndFor
\State Construct the binary adjacency matrix $G$: set $G_{ij}=1$ if
  $i$ selects $j$ or $j$ selects $i$, and $G_{ij}=0$ otherwise
\State Normalize the graph affinities: $p_{ij} \gets G_{ij}/\sum_{u,v}G_{uv}$

\Statex \textit{Initialization and optimization}
\State Form $Y_0$ from columns 2 and 3 of $E$
  \Comment{skip the constant first mode}
\State Fix column signs, rescale $Y_0$, and add Gaussian jitter using seed $s$
\State \Return $\Call{tSNE}{p,Y_0,s}$
\end{algorithmic}
\end{algorithm}

The self-OT, Gram, kNN, and zero variants use the same cross-batch
plan $P$ and replace the two within-batch blocks of $A$ as described
in Section~\ref{sec:method}.

\textbf{Affinity construction.} We used k=15 neighbors, a commonly used neighborhood size in nearest-neighbor graph methods and the default in UMAP and NCVis~\citep{mcinnes2018umap,artemenkov2020ncvis}.  We set the entropic regularization hyperparameter to the median of the Euclidean cross-batch cost, following \citet{landa2024entropic}. Therefore, relative regularization hyperparameter $r=1$ by default. We computed the transport plan using Sinkhorn iterations with a convergence tolerance of \(10^{-12}\) and a maximum of \(10^4\) iterations. 

For spectral within-batch affinities, any negative entries are explicitly set to zero. At the default regularization, they are rare and extremely small: all datasets except nCoV do not have negative entries, while nCoV has fewer than 30, with the largest $-2.1 \cdot 10^{-8}$. 

\textbf{Initialization}
The t-SNE optimization requires an initial 2D layout. We considered four initializations.
(1) Random initialization draws each coordinate independently from a zero-mean Gaussian with standard deviation $10^{-4}$, following \texttt{openTSNE}~\citep{polivcar2024opentsne}.
(2) As in openTSNE, spectral initialization computes a diffusion map of the OT-tSNE affinity graph using the two leading nontrivial eigenvectors of its random-walk transition matrix, each weighted by its eigenvalue.
(3) PCA initialization projects the concatenated batches onto their first two joint principal components.
(4) EOT initialization uses the second and third left singular vectors of the cross-batch transport plan for the first batch and the corresponding right singular vectors for the second.

We use EOT initialization by default. For the spectral, PCA, and EOT initializations, we rescale both coordinates by a common factor so that the first has standard deviation $10^{-4}$. 

By default, openTSNE adds a small jitter to deterministic initializations such as PCA, so that no two points sit on the same spot. Therefore, following openTSNE, by default, we add independent Gaussian noise with standard deviation $10^{-6}$ to each coordinate to separate coincident points. We use the run's random seed to generate the noise.

\textbf{t-SNE parameters.}
We supplied the sparse OT-tSNE affinity matrix and initial layout to \texttt{openTSNE} and used its default settings for all optimization parameters. These include 250 iterations with early exaggeration followed by 500 iterations without exaggeration.

\subsection{Benchmark setup}

\textbf{t-SNE parameters for two-step embedding methods.}
For general-purpose integration methods that first construct a 50-dimensional representation and subsequently use t-SNE for visualization, we applied the default openTSNE configuration. Specifically, EOT 50D, PCA 50D, Harmony 50D, Scanorama 50D, scVI 50D, Seurat 50D, and LIGER 50D were embedded with perplexity 30, PCA initialization, 250 early-exaggeration iterations, and 500 post-exaggeration iterations. 

\textbf{Benchmarked methods.}
All competitors (EOT, Harmony, Scanorama, scVI, BC t-SNE, RC t-SNE, MultiMAP, Seurat, LIGER, and Ref t-SNE) used their default hyperparameters, except for the specified target dimensionalities (2 and 50) and random seed. Additionally, Seurat, LIGER, and Ref t-SNE required modifications to their standard input preprocessing to ensure consistent feature selection across methods, as detailed below. 

\textbf{Input formats.}
Different methods require different input representations. However, to minimize differences due to preprocessing and enable a more direct comparison of the integration methods, all methods used the same cells and the same set of $1\,000$ selected genes. 
\begin{itemize}
    \item OT-tSNE, EOT, joint t-SNE, Harmony, and RC t-SNE received the batches in the 50-dimensional PCA representation computed from the log-normalized $1\,000$-gene features. 
    \item PCA, Scanorama, MultiMAP, and BC t-SNE received the log-normalized $1\,000$-gene representation, as these methods work directly in feature space.
    \item scVI, Seurat, LIGER, and Ref t-SNE normally receive raw counts and perform additional preprocessing internally. To maintain the same feature set across methods, we provided raw counts restricted to the common $1\,000$-gene panel. For Seurat, LIGER, and Ref t-SNE, we therefore disabled their internal gene-selection procedures while retaining the remainder of their standard workflows.
\end{itemize}

\textbf{Computational environment.}
All methods were run on the same computing cluster with Intel Xeon Gold 6240 processors, with four CPUs allocated per run and no GPU. Thread counts were set to four where supported. For some benchmarked methods, some stages (MultiMAP layout optimization, Scanorama's exact neighbor search, and BC t-SNE optimization) used fewer cores due to parallelization limitations in the original implementations. We report median runtimes over three seeds per dataset. Timing covers the algorithm, excluding data loading, result writing, metric evaluation, and interpreter startup and imports. For 50-dimensional integrations, runtimes include the subsequent t-SNE visualization. Shared PCA preprocessing is excluded for methods receiving principal components, whereas dimensionality reduction performed internally by methods receiving gene expression is included.

\subsection{Hyperparameters of the compared methods}
\label{app:hyperparameter_comparison}

\begin{table}[tb]
    \centering
    \small
    \setlength{\tabcolsep}{4pt}
    \renewcommand{\arraystretch}{1.15}
    \caption{Selected hyperparameters of the compared methods, with software defaults in parentheses unless otherwise stated. The last column specifies changes to the listed settings in our benchmark.$B$ denotes the number of batches and $N$ the number of cells.}
    \label{tab:hyperparameters}

    \begin{tabularx}{\linewidth}{
        @{}
        >{\raggedright\arraybackslash}p{0.17\linewidth}
        >{\raggedright\arraybackslash}X
        >{\raggedright\arraybackslash}p{0.22\linewidth}
        @{}
    }
        \toprule
        Method & Selected hyperparameters & Benchmark changes \\
        \midrule

        \multicolumn{3}{@{}l}{
            \textit{Integrated visualization methods}
        } \\
        \addlinespace

        OT-tSNE &
        Neighborhood size (15),
        relative entropic regularization (1) &
        --- \\
        \addlinespace

        MultiMAP &
        Neighborhood size ($15B$),
        dataset weights (0.5 each for two batches),
        minimum embedding distance (0.25),
        embedding spread (1) &
        --- \\
        \addlinespace

        RC t-SNE &
        Same-batch attraction factor
        ($10^{-1}$),
        perplexity (30) &
        Used the recommended attraction factor $10^{-4}$ \\
        \addlinespace

        BC t-SNE &
        Batch-corrected projection rank (50),
        perplexity (30) &
        --- \\
        \addlinespace

        Reference t-SNE &
        Reference perplexities (50 and 500),
        query perplexity (30),
        distance measure (cosine) &
        --- \\

        \midrule
        \multicolumn{3}{@{}l}{
            \textit{General-purpose integration methods}
        } \\
        \addlinespace

        Seurat v5 (CCA) &
        CCA dimensions (30),
        anchor-finding neighborhood (5),
        anchor filtering (disabled),
        anchor-scoring neighborhood (30),
        anchor-weighting neighborhood (100),
        anchor-weighting kernel width (1) &
        CCA dimensions: 2 for 2D, default for 50D \\
        \addlinespace

        LIGER &
        Number of factors (20),
        penalty on dataset-specific factors (5),
        quantile-normalization neighborhood (20),
        number of quantiles (50) &
        Factors: 2 or 50 \\
        \addlinespace

        Scanorama &
        Matching neighborhood (20),
        merging threshold (0.1),
        reduced dimension (100),
        correction kernel width (15) &
        Reduced dimension: 2 or 50 \\
        \addlinespace

        scVI &
        Latent dimension (10),
        hidden-layer width (128),
        number of hidden layers (1),
        dropout rate (0.1),
        count likelihood (zero-inflated negative binomial),
        dispersion model (per gene) &
        Latent dimension: 2 or 50 \\
        \addlinespace

        Harmony &
        Diversity penalty (2),
        ridge penalty (estimated from the data),
        number of clusters ($\min(N/30,100)$),
        soft-assignment width (0.1) &
        --- \\
        \addlinespace

        EOT &
        Relative entropic regularization (1),
        number of singular modes (2),
        diffusion time (0) &
        Singular modes: 2 or 50 \\

        \bottomrule
    \end{tabularx}
\end{table}

We summarize selected hyperparameters governing affinity construction, batch correction, and representation learning in the compared methods (Tab.~\ref{tab:hyperparameters}). These include neighborhood sizes, regularization strengths, representation dimensions, and model architecture. The table provides a descriptive overview rather than an exhaustive list or a measure of the tuning effort required by each method. We omit optimization and numerical settings, shared preprocessing, and hyperparameters of the subsequent t-SNE visualization of 50-dimensional integrations. We evaluated all methods without dataset-specific hyperparameter tuning.
\section{Evaluation metrics}
\label{app:metrics}

\paragraph{kBET.}
k-nearest neighbor batch effect test~\citep[kBET]{buttner2019test} tests whether a cell's neighborhood has the expected batch composition. We evaluate each cell type separately and retain types with at least 45 cells in each batch. For every retained cell, we find its 15 nearest neighbors within the same type, excluding the query cell, and count from which batches they are. The expected count for each batch is 15 times its proportion within that cell type. We compare the observed batch counts with the expected counts using a Pearson chi-square test with one degree of freedom, since our datasets contain two batches. We calculate the fraction of tests with $p\geq0.05$ within each type, then average these fractions equally across types. A higher score means fewer neighborhoods show a significant difference from the expected batch composition.

\paragraph{iLISI.}
Integration local inverse Simpson's index~\citep[iLISI]{korsunsky2019harmony} measures the effective number of batches in a cell's neighborhood. We use $K=90$ nearest neighbors in the joint embedding with a perplexity of 30. Let $\mathcal N_K(i)$ denote these neighbors, excluding cell $i$, and let $d_{ij}$ be the Euclidean distance between cells
$i$ and $j$. We compute neighbor weights and local batch proportions as
\begin{equation}
 w_{ij}=
 \frac{\exp(-\beta_i d_{ij})}
 {\sum_{r\in\mathcal N_K(i)}\exp(-\beta_i d_{ir})},
 \qquad
 q_{ib}=\sum_{j\in\mathcal N_K(i)}
 w_{ij}\mathbf{1}\{b_j=b\},
\end{equation}
where $b_j$ is the batch of cell $j$, and $\beta_i$
is adjusted to achieve the target perplexity. With $B=2$ batches, we compute
\begin{equation}
 L_i=\left(\sum_{b=1}^{B}q_{ib}^{2}\right)^{-1},
 \qquad
 S_{\mathrm{iLISI}}=
 \frac{\operatorname{median}_{i}L_i-1}{B-1}.
\end{equation}
As in the original iLISI implementation, the weights use unsquared Euclidean distances.

\paragraph{Neighborhood purity.}
Neighborhood purity measures the fraction of neighbors that share a cell's type. Let $I_c$ be the set of $n_c$ cells of type $c$. We evaluate  types with at least five cells, $\mathcal C=\{c:n_c\geq5\}$, using $k_c=\min(15,n_c-1)$ neighbors:
\begin{equation}
 S_{\mathrm{purity}}=
 \frac{1}{|\mathcal C|}\sum_{c\in\mathcal C}
 \frac{1}{n_c}\sum_{i\in I_c}
 \frac{1}{k_c}\sum_{j\in\mathcal N_{k_c}(i)}
 \mathbf{1}\{c_j=c\}.
\end{equation}
Here, $c_j$ is the type of cell $j$. Neighbors are selected from the entire embedding, excluding the query cell. Cells from types with fewer than five cells remain candidate neighbors. We average first within each evaluated type and then across types, giving each type equal weight.

\paragraph{Silhouette score.}
The silhouette score measures cell-type separation \citep{rousseeuw1987silhouettes}. For cell $i$, let $a_i$ be its mean Euclidean distance to other cells of the same type. Let $h_i$ be the smallest of its mean Euclidean distances to each of the other types. We use scikit-learn to compute silhouette
widths $s_i\in[-1,1]$ for all $n$ cells and rescale their mean to $[0,1]$:
\begin{equation}
 s_i=\frac{h_i-a_i}{\max(a_i,h_i)},
 \qquad
 S_{\mathrm{ASW}}=
 \frac{1}{2}\left(\frac{1}{n}\sum_{i=1}^{n}s_i+1\right).
\end{equation}
Each cell has equal weight, so larger cell types contribute more to the final score.

\subsection{Ranking procedure}
\label{app:ranking}
For each dataset and regularization setting, we first average every metric across random seeds. Higher values indicate better performance for all four scores. We min--max normalize each metric across methods,
\begin{equation*}
\tilde m_i=\frac{m_i-\min_j m_j}{\max_j m_j-\min_j m_j}.
\end{equation*}
Within each dataset, each method gets an overall score $S_{\mathrm{overall}}$, defined as the average over all four normalized metrics. Then we rank the methods by $S_{\mathrm{overall}}$ within each dataset, with rank 1 denoting the best method. The final aggregate rank of a method is the mean of its dataset-level ranks (Tab.~\ref{tab:ranks}). 

We also average the normalized metrics within the biological conservation and batch mixing groups to obtain scores $S_{\mathrm{bio}}$ and $S_{\mathrm{mix}}$, respectively. For the top-3 finish ranking (Fig.~\ref{fig:fig1}g), we rank the methods separately by $S_{\mathrm{bio}}$ and $S_{\mathrm{mix}}$ within each dataset, and we count how often each method attains a top-3 rank across the seven datasets and these two scores.

\begin{figure}[h]
    \centering
    \includegraphics[width=\linewidth]{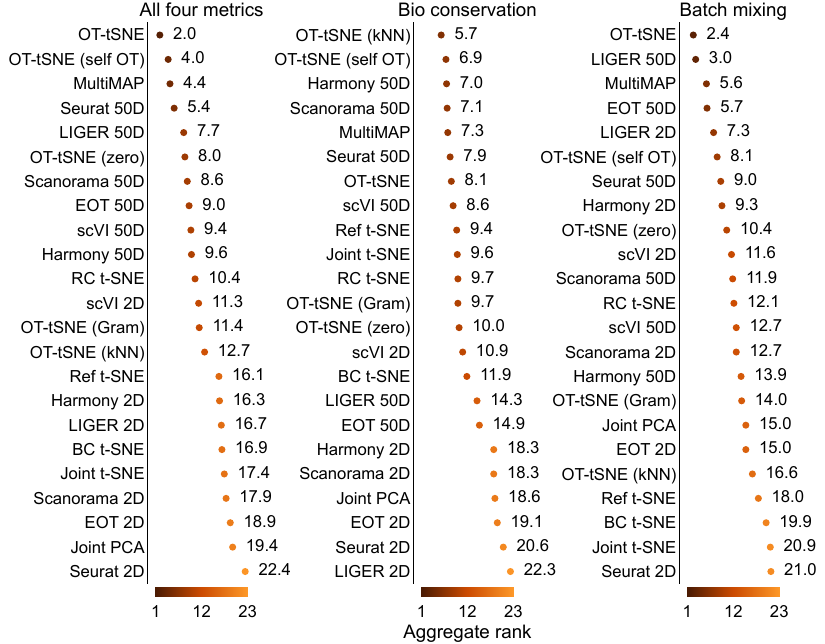}
    \caption{Mean rank across seven datasets. Within each dataset, we ranked methods by the average of all four normalized metrics (left), the two biological conservation metrics (middle), or the two batch mixing metrics (right). Lower ranks indicate better performance.}
    \label{app:fig:ranks}
\end{figure}

For $S_{\mathrm{bio}}$ and $S_{\mathrm{mix}}$, we likewise averaged the within-dataset ranks across the seven datasets (Fig.~\ref{app:fig:ranks}). These ranks summarize biological conservation and batch mixing separately. A method that excels at biological conservation may rank lower overall if it mixes batches poorly, and vice versa. A method that performs well on both objectives can therefore rank highly overall even with few top-three finishes in either category. For example, OT-tSNE (self OT) ranked second overall with a mean rank of 4.0 (Fig.~\ref{app:fig:ranks}), yet its only category-specific top-three finish was for biological conservation on Mouse Atlas (Fig.~\ref{fig:fig1}g). 

\section{Sensitivity analysis}

\subsection{Sensitivity to neighborhood size and entropic regularization}
\label{app:hyperparameter_sensitivity}

We evaluated OT-tSNE with random initialization on 25 combinations of neighborhood size $k\in\{5,10,15,30,60\}$ and relative regularization $r\in\{0.1,0.5,1,5,10\}$ for 3 random seeds across the seven real datasets. For each dataset, we averaged each metric across seeds and measured its change from the default setting $(k,r)=(15,1)$. We then averaged these changes across datasets. 

The overall score changed little near the default, although individual metrics showed a trade-off between batch mixing and biological conservation. At fixed $k$, weaker regularization generally favored batch mixing, particularly iLISI, at the expense of biological conservation. At $r=1$, increasing $k$ tended to improve biological conservation, with smaller average gains beyond $k=15$ (Fig.~\ref{app:fig:hyperparameter_lines}). These trade-offs largely offset each other near the default setting. 

\begin{figure}[t]     
    \centering     
    \includegraphics[width=\linewidth]{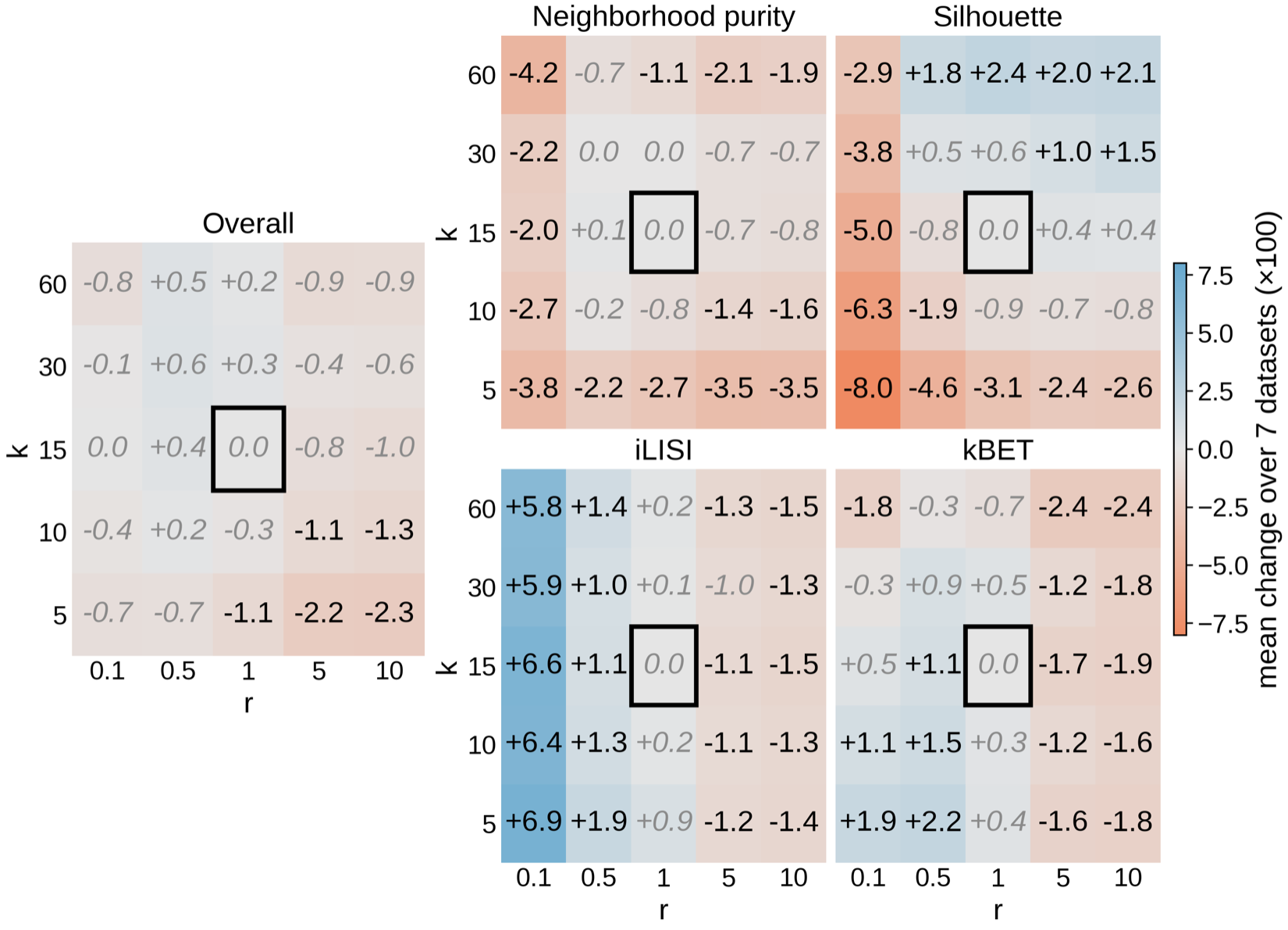}     
    \caption{Sensitivity of OT-tSNE to neighborhood size $k$ and relative regularization $r$. Values show the mean change from the boxed default setting $(k,r)=(15,1)$ across seven datasets, multiplied by 100. The overall score is the mean of the four metrics. Black labels indicate changes of at least 0.01 in absolute value on the original scale that also exceed twice their estimated standard error.} 
    \label{app:fig:hyperparameter_pooled} 
\end{figure}

\begin{figure}[p]     
    \centering    
    \includegraphics[width=\linewidth,height=0.83\textheight,keepaspectratio]{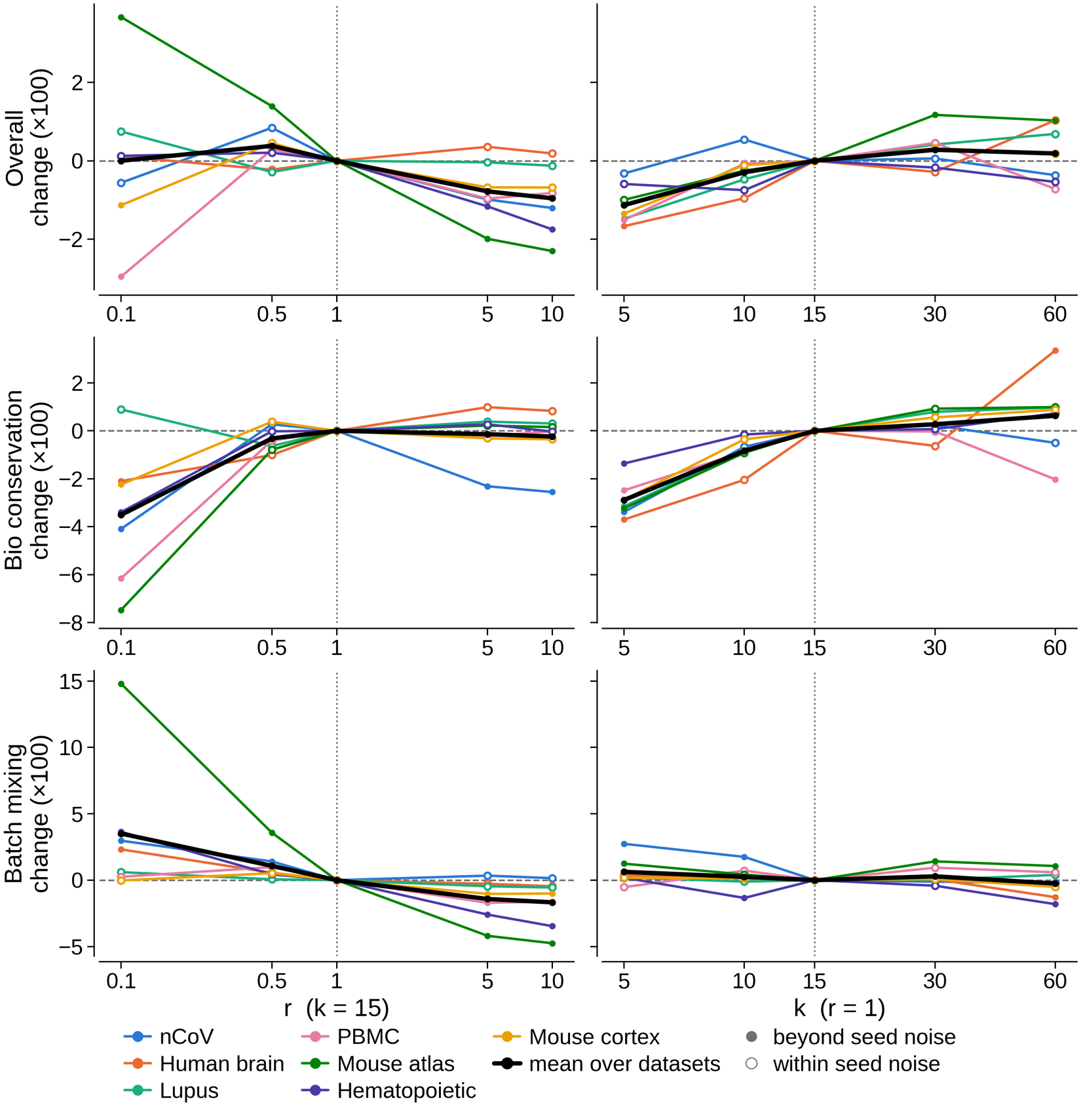}     
    \caption{Dataset-level changes when varying $r$ at $k=15$ (left) or $k$ at $r=1$ (right). From top to bottom, rows show the overall score (mean of all four metrics), biological conservation (mean of neighborhood purity and silhouette), and batch mixing (mean of iLISI and kBET). Changes are relative to the default setting and multiplied by 100. Colored lines show individual datasets and the black line shows their mean. Filled and open colored markers indicate whether a change met both thresholds defined in Fig.~\ref{app:fig:hyperparameter_pooled}.}     
    \label{app:fig:hyperparameter_lines} 
\end{figure}

Across the full grid, no alternative improved the mean overall score by even one point on the $\times 100$ scale (Fig.~\ref{app:fig:hyperparameter_pooled}). We therefore retained the literature-based defaults $(k,r)=(15,1)$, which provided a stable balance between batch mixing and biological conservation. Competing methods were likewise evaluated without dataset-specific hyperparameter tuning, using the configurations described in Appendix~\ref{app:hyperparameter_comparison}.

\clearpage
\subsection{Sensitivity to initialization}
\label{app:initialization_sensitivity}

We compared random, PCA, spectral, and EOT initialization for all five OT-tSNE variants on the seven real datasets. For each dataset and variant, we averaged each metric over three seeds and computed its difference from the random-initialization baseline. We then averaged these differences across datasets.

\begin{figure}[tb]
    \centering
    \includegraphics[width=\linewidth]{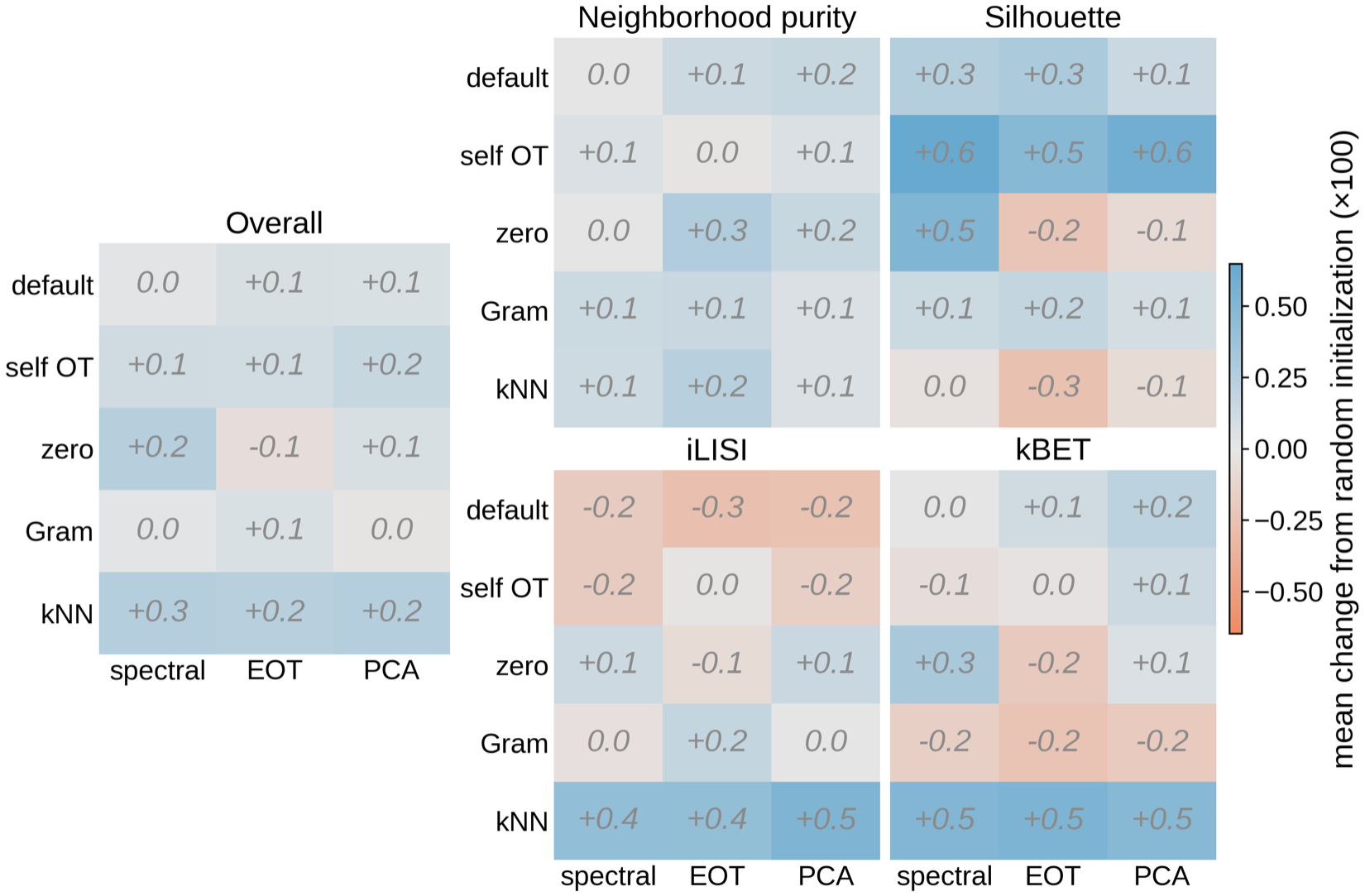}
    \caption{Mean change from random initialization across seven datasets, multiplied by 100. Each panel shows one metric and the mean of the four metrics, with OT-tSNE variants in rows and alternative initializations in columns. All values fall below the magnitude threshold of 1.}
    \label{app:fig:init_pooled}
\end{figure}

\begin{figure}[p]
    \centering
    \includegraphics[width=\linewidth,height=0.84\textheight,keepaspectratio]
        {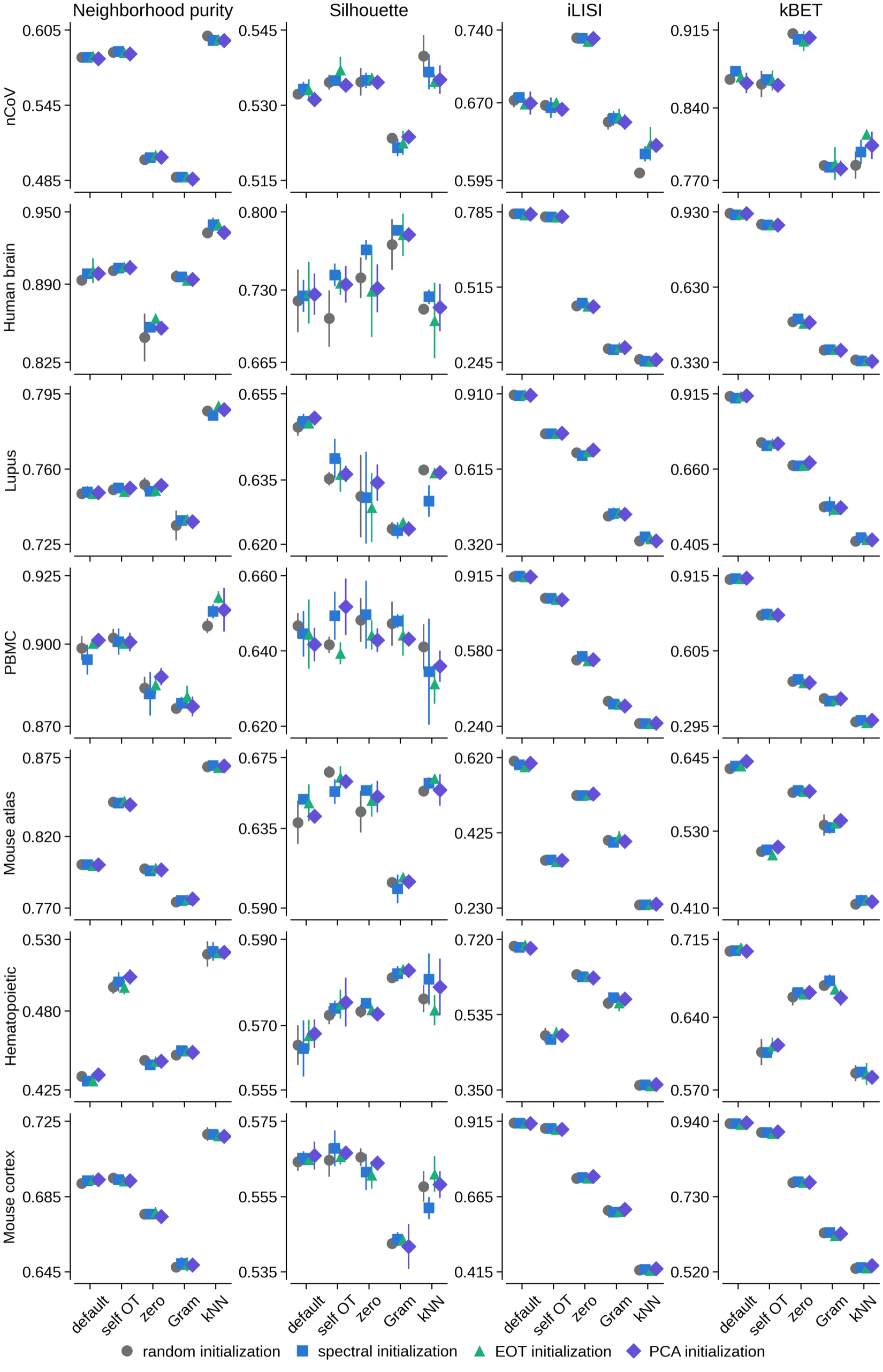}
    \caption{Effect of initialization on individual datasets. Each row corresponds to one dataset, and each column corresponds to one metric. For each OT-tSNE variant, mean scores over three seeds for random, spectral, EOT, and PCA initialization are displayed. }
    \label{app:fig:init_datasets}
\end{figure}

The pooled changes were small for all four metrics and all five variants (Fig.~\ref{app:fig:init_pooled}). The largest absolute change was approximately 0.0065 in silhouette score for the self-OT variant. Some differences were larger on individual datasets (Fig.~\ref{app:fig:init_datasets}).

\begin{figure}[p]
    \centering
    \includegraphics[width=\linewidth,height=0.84\textheight,keepaspectratio]
        {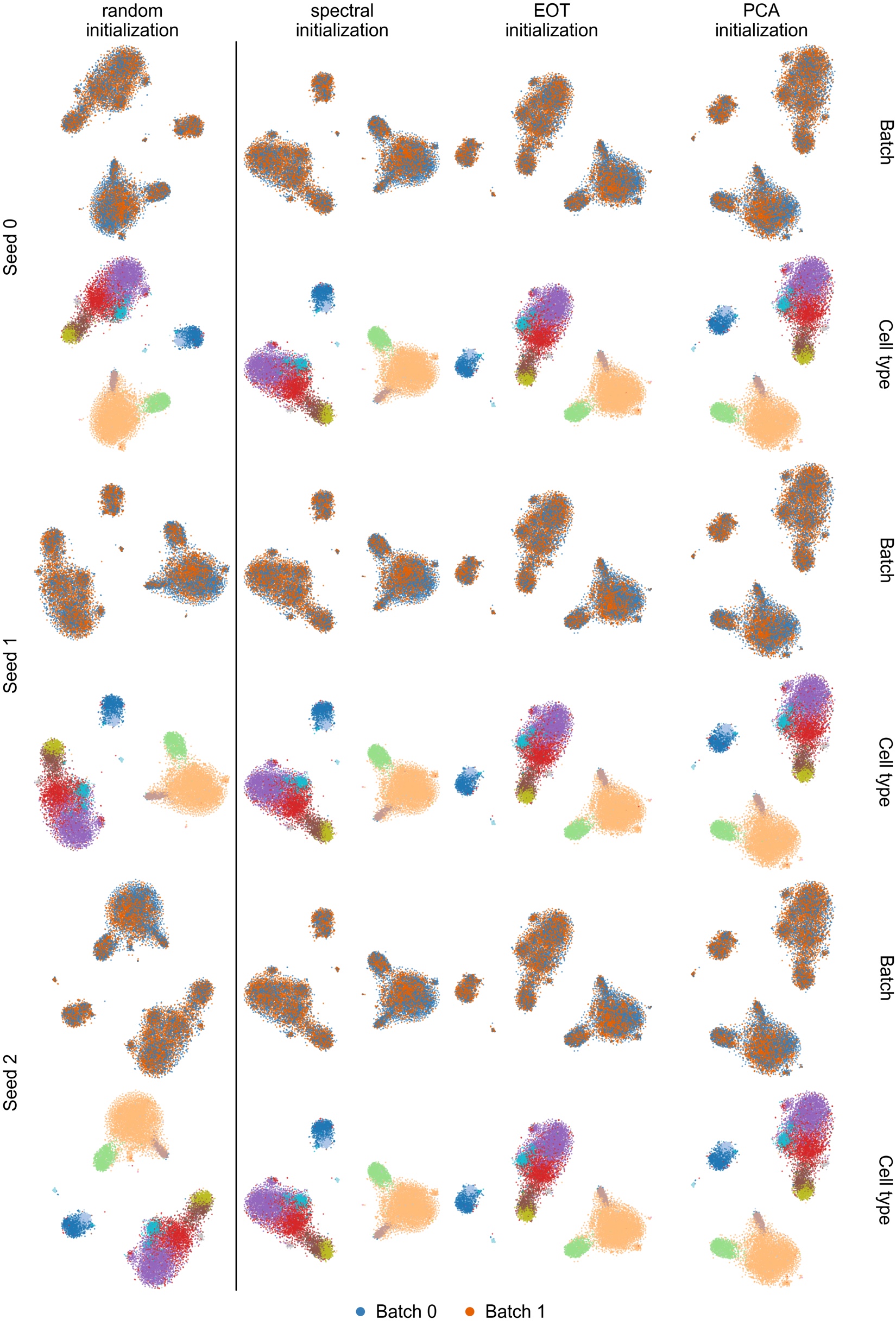}
    \caption{OT-tSNE embeddings of the Lupus dataset across three seeds and four initializations. Columns show initialization methods. Each pair of rows shows the embeddings for one seed, colored by batch (top) and cell type (bottom). With jitter disabled, spectral, EOT, and PCA initialization each reproduced its own layout across seeds. With random initialization, the relative positions of the major clusters varied.}
    \label{app:fig:init_seeds_lupus_nojitter}
\end{figure}

\paragraph{Stability of the embedding across seeds.}
Because OT-tSNE is designed for visual exploration, we examined whether the same cell-type clusters appeared across runs and whether their relative positions were reproducible. We compared Lupus embeddings across three seeds. OT-tSNE constructs its affinity matrix deterministically, and given the same affinities and initial coordinates, the \texttt{openTSNE} optimizer is deterministic in the same computing environment. For this comparison, we disabled the Gaussian jitter normally added to the spectral, EOT, and PCA initializations. Each of these initializations then reproduced the same layout across all three seeds (Fig.~\ref{app:fig:init_seeds_lupus_nojitter}). With random initialization, the major clusters remained recognizable by their cell-type composition, although their relative positions varied across seeds.

\clearpage
\section{Computational complexity}
\label{app:comp_complexity}
\subsection{OT-tSNE}

Let $n_0,n_1$ be the batch sizes, $N=n_0+n_1$ the total number of cells, $d$ the input dimension, and $m=\min(n_0,n_1)$. We analyze the dense transport plan and spectral within-batch construction used by default.
Tab.~\ref{tab:ot_complexity} summarizes the costs before t-SNE optimization. Here, $T$ is the number of Sinkhorn iterations, $q$ the retained SVD rank, $p$ the oversampling parameter, $s$ the number of power iterations, and $\ell=q+p\leq m$.

\begin{table}[tb]
    \centering
    \vspace*{-3mm}
    \small
    \setlength{\tabcolsep}{4pt}
    \renewcommand{\arraystretch}{1.15}
    \caption{Time complexity of OT-tSNE affinity construction using full
    or randomized SVD. The last column gives the scaling for comparable
    batch sizes, holding $d,T,k,q,p,s$ fixed in the truncated case.
    Full reconstruction uses all $m$ singular directions.}
    \label{tab:ot_complexity}
    \begin{tabularx}{\linewidth}{
        @{}>{\raggedright\arraybackslash}X l c@{}
    }
        \toprule
        Step & Time complexity & $n_0\approx n_1$ \\
        \midrule
        Distances and median
        & $O(n_0n_1d)$ & $O(N^2)$ \\

        Sinkhorn iterations
        & $O(Tn_0n_1)$ & $O(N^2)$ \\

        Full SVD
        & $O(n_0n_1m)$ & $O(N^3)$ \\

        Randomized SVD
        & $O((s+1)[n_0n_1\ell+N\ell^2])$ & $O(N^2)$ \\

        Full reconstruction
        & $O((n_0^2+n_1^2)m)$ & $O(N^3)$ \\

        Truncated reconstruction
        & $O((n_0^2+n_1^2)q)$ & $O(N^2)$ \\

        Sparsification
        & $O(N^2)$ & $O(N^2)$ \\
        \bottomrule
    \end{tabularx}
\end{table}

\paragraph{Transport and spectral construction.}
Each Sinkhorn iteration involves matrix--vector products with the dense cross-batch kernel~\citep{cuturi2013sinkhorn}. The number of iterations depends on the data, regularization, and convergence tolerance; at our default regularization, convergence required only 11--21 iterations across the benchmarked datasets. For comparable batch sizes, both computing the full SVD of the transport plan and constructing the within-batch affinity matrices from all singular directions have cubic cost in $N$. Randomized SVD computes an approximate decomposition using fewer singular directions~\citep{halko2011finding}. Truncated SVD reduces affinity construction to quadratic cost for fixed approximation rank and other parameters. 

\paragraph{Sparsification and optimization.}
We retain the largest affinities in each block using top-$k$ selection, which costs $O(N^2)$ for fixed $k$. Symmetrizing the selected edges leaves $O(Nk)$ nonzero affinities. The attractive forces therefore cost $O(Nk)$ per t-SNE iteration. We use the default automatic selection in \texttt{openTSNE}, which evaluates repulsive forces using Barnes--Hut for $N<10{,}000$ and FFT interpolation for larger datasets~\citep{polivcar2024opentsne}. Barnes--Hut approximates the repulsive forces with a typical cost of $O(N\log N)$ per iteration~\citep{maaten2014accelerating}. FFT interpolation has linear dependence on $N$ when the interpolation grid and order are fixed~\citep{linderman2019fast}. These per-iteration costs are multiplied by the number of optimization iterations.

\paragraph{Memory and practical runtime.}
Affinity construction requires $O(N^2)$ memory because it stores dense cross-batch costs and transport weights, as well as dense within-batch affinities. Truncated SVD reduces the storage needed for singular vectors, but these dense matrices remain. After sparsification, the graph requires $O(Nk)$ storage. Barnes--Hut t-SNE has linear memory requirements for fixed neighborhood size~\citep{maaten2014accelerating}. FFT interpolation additionally stores the interpolation grid and associated working arrays~\citep{linderman2019fast}. 

Fig.~\ref{app:fig:runtime} compares full SVD with randomized SVD truncated to 200 components. This approximation reduced total runtime by up to a factor of five, while seed-averaged metric scores changed by at most $0.012$ in absolute value across the benchmarked datasets. The other OT-tSNE variants avoid SVD and were also faster than the full-SVD OT-tSNE (Fig.~\ref{app:fig:runtime}).

\subsection{Benchmarked methods}

Tab.~\ref{tab:competitor_complexity} broadly summarizes the main computational
steps of the compared methods, published scaling results, and
operation counts for the benchmarked implementations. Since this summary concerns
specific steps or experimental settings, Fig.~\ref{app:fig:runtime} compares measured runtimes under our benchmark settings.

\begin{table}[tb]
    \centering
    \small
    \setlength{\tabcolsep}{4pt}
    \renewcommand{\arraystretch}{1.15}
    \caption{Main computational steps of the benchmarked methods with costs. $N$ denotes the number of cells, $k$ the neighborhood size, and $G$ the number of interpolation-grid points.
    The notation $n_0,n_1,d,T$ follows the preceding subsection.}
    \label{tab:competitor_complexity}
    \begin{tabularx}{\linewidth}{
        @{}
        >{\raggedright\arraybackslash}p{0.13\linewidth}
        >{\raggedright\arraybackslash}X
        >{\raggedright\arraybackslash}X
        @{}
    }
        \toprule
        Method & Main computation & Cost and assumptions \\
        \midrule

        MultiMAP &
        Find neighbors within and across datasets. &
        The published $O(N^{1.14})$ estimate relies on empirical
        scaling of approximate neighbor search
        ~\citep{jain2021multimap}. \\
        &
        Optimize a UMAP-based objective on the MultiMAP graph
        using negative sampling. &
        $O(Nk)$ per epoch for a fixed number of negative
        samples per edge~\citep{jain2021multimap}. \\
        \addlinespace

        RC t-SNE &
        Optimize the embedding using FFT-accelerated t-SNE. &
        Combining sparse attraction and FFT repulsion gives
        $O(Nk+N+G\log G)$ per iteration for fixed interpolation
        order. This excludes affinity construction
        ~\citep{heiter2023rctsne,linderman2019fast}. \\
        \addlinespace

        Ref t-SNE &
        Build and optimize a reference embedding. &
        Uses accelerated t-SNE, with the optimization costs
        described above
        ~\citep{policar2023embedding,polivcar2024opentsne}. \\
        &
        Place query cells into the fixed reference embedding. &
        $O(M)$ per iteration for $M$ query cells with fixed
        neighborhood size and FFT interpolation grid.
        This excludes query neighbor search
        ~\citep{policar2023embedding,polivcar2024opentsne}. \\
        \addlinespace

        BC t-SNE &
        Compute a batch-corrected projection, then apply t-SNE. &
        The projection cost is described as comparable to
        PCA~\citep{aliverti2020projected}.
        The benchmarked implementation evaluates the exact
        t-SNE gradient on a single thread, costing $O(N^2)$
        per iteration. \\
        \midrule

        Seurat  &
        Compute CCA, find anchors, and correct the data. &
        In the benchmarked implementation (Seurat 5.5.1),
        this matrix requires $O(n_0n_1)$ memory. \\
        \addlinespace

        LIGER &
        Jointly factorize the datasets, then align the factors
        by quantile normalization. &
        Factorization time was empirically linear in $N$
        at fixed gene count and rank. The reported scaling
        concerns factorization, not the subsequent alignment
        ~\citep{liu2020jointly}. \\
        \addlinespace

        Scanorama &
        Reduce dimensionality using randomized SVD. &
        $O(Ng)$ for $g$ genes when the retained rank,
        oversampling, and number of power iterations are fixed
        ~\citep{hie2019scanorama,halko2011finding}. \\
        &
        Find cross-batch neighbors and identify matching cells. &
        The published approach accelerates neighbor search
        using locality-sensitive hashing
        ~\citep{hie2019scanorama}. \\
        &
        Correct the data using matched cell pairs. &
        $O(Nga)$ per merge for $g$ corrected
        dimensions, and $a$ matched pairs
        ~\citep{hie2019scanorama}. \\
        \addlinespace

        scVI &
        Train a variational autoencoder using minibatches. &
        For fixed network architecture and gene count, the cost is $O(N)$
        per epoch ~\citep{lopez2018scvi}. \\
        \addlinespace

        Harmony &
        Alternate between clustering and batch correction. &
        Runtime depends on the number of clusters and
        optimization iterations
        ~\citep{korsunsky2019harmony}. \\
        \addlinespace

        EOT &
        Compute cross-batch distances and the transport plan. &
        $O(n_0n_1(d+T))$ for distance computation and $T$
        Sinkhorn iterations, as in OT-tSNE
        ~\citep{cuturi2013sinkhorn,landa2024entropic}. \\
        &
        Compute the leading singular vectors. &
        $O(n_0n_1)$ \\
        \bottomrule
    \end{tabularx}
\end{table}

\clearpage
\section{Additional figures}
\label{app:real_benchmark}

\begin{figure}[h]
    \centering
    \includegraphics[width=\linewidth]{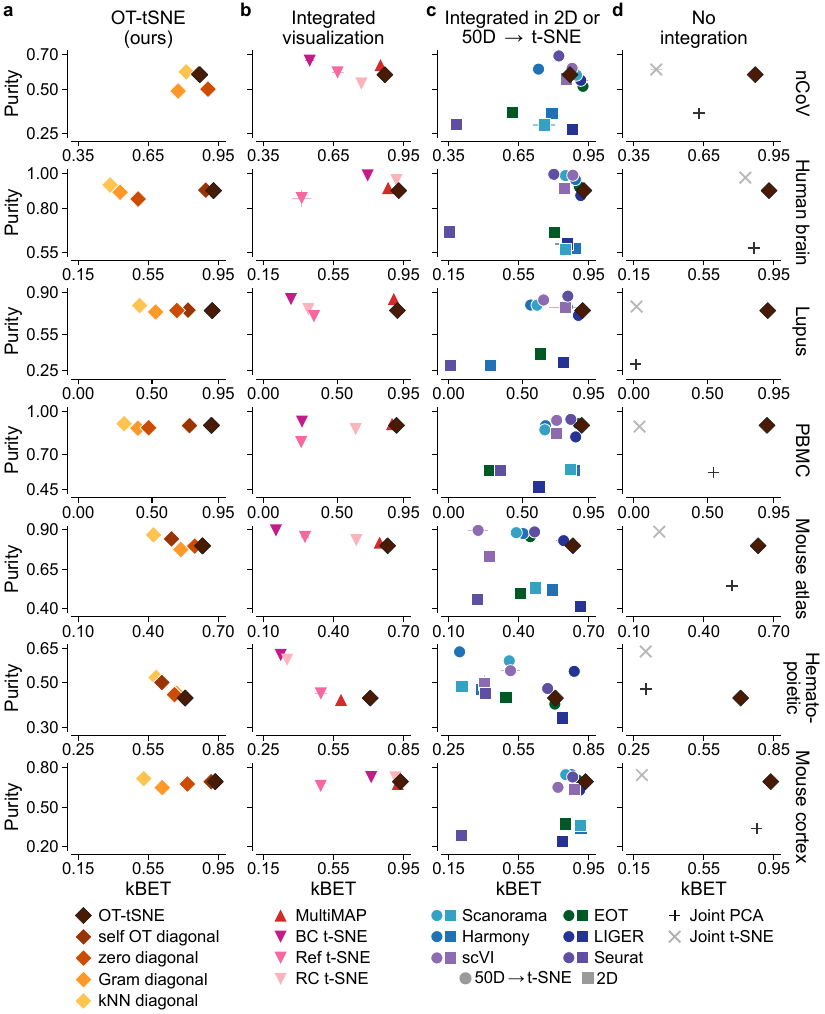}
    \caption{Benchmark results for all methods and datasets for the purity and kBET metrics for \textbf{a.} OT-tSNE variants, \textbf{b.} integrated visualization methods, \textbf{c.} general-purpose integration methods either with 50D latent spaces visualized with t-SNE (dots) or with 2D latent spaces directly used for visualization (squares), and \textbf{d.} baselines without integration. Reported means and standard deviations over datasets. OT-tSNE is repeated in all panels for better comparability.}
    \label{app:fig:benchmark_dataset_kbet_purity}
\end{figure}

\begin{figure}[tb]
    \centering
    \includegraphics[width=\linewidth]{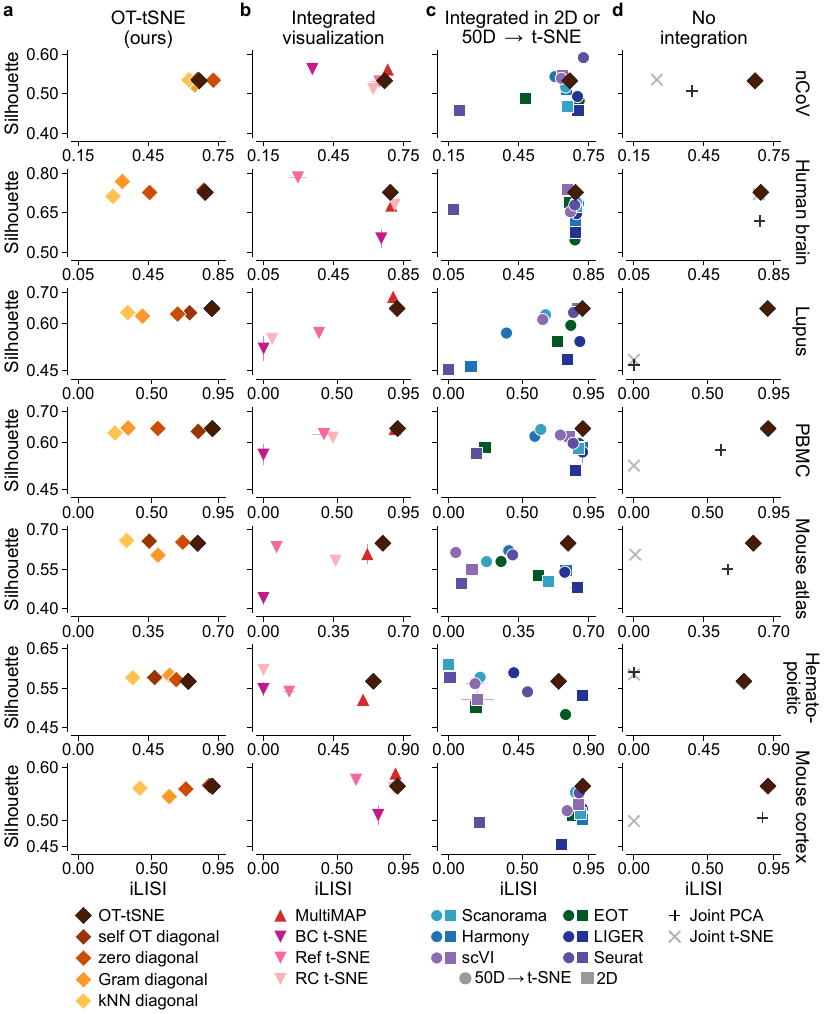}
    \caption{Benchmark results for all methods and datasets for the silhouette and iLISI metrics for \textbf{a.} OT-tSNE variants, \textbf{b.} integrated visualization methods, \textbf{c.} general-purpose integration methods either with 50D latent spaces visualized with t-SNE (dots) or with 2D latent spaces directly used for visualization (squares), and \textbf{d.} baselines without integration. Reported means and standard deviations over datasets. OT-tSNE is repeated in all panels for better comparability.}
    \label{app:fig:benchmark_dataset_ilisi_silhouette}
\end{figure}

\begin{figure}[tb]
    \centering
    \includegraphics[width=0.95\linewidth]{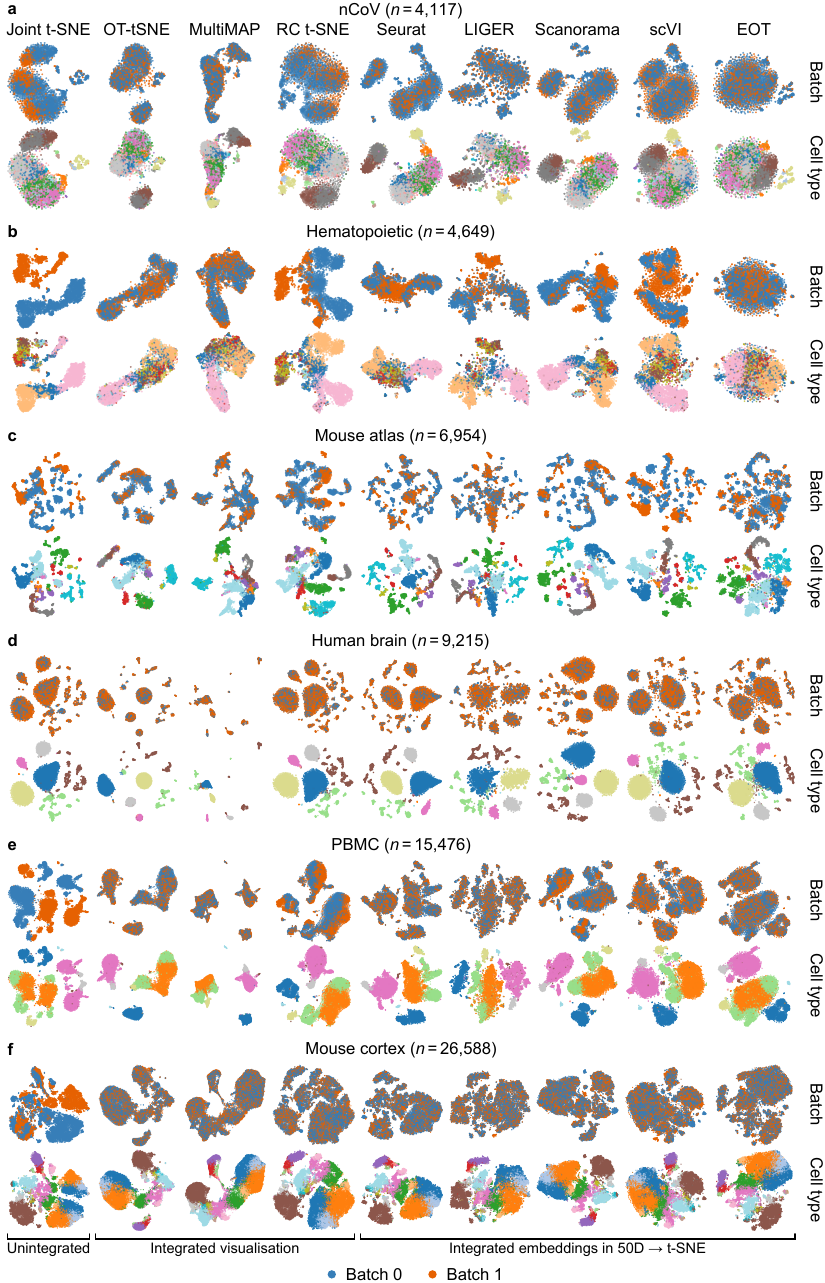}
    \caption{Visualizations of selected methods on the remaining 6 datasets, not shown in Fig.~\ref{fig:embeddings_lupus}. Each embedding is shown twice, in the top row colored by batch, in the bottom row colored by cell type.}
    \label{app:fig:embeddings_other_datasets}
\end{figure}

\begin{figure}[tb]
    \centering
    \includegraphics[width=0.89\linewidth]{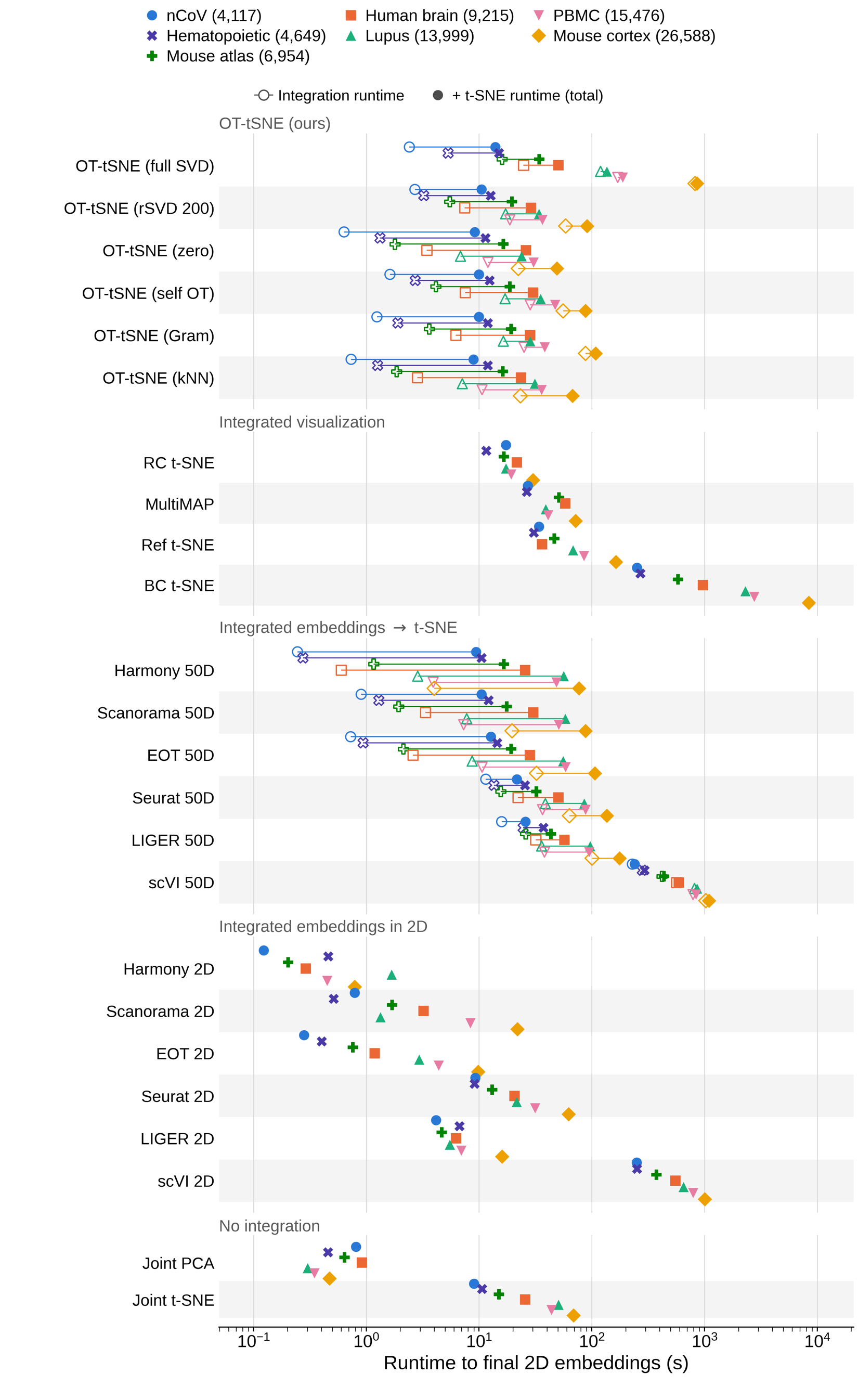}
    \caption{Median runtimes of all methods on all datasets over three random seeds. For OT-tSNE variants, we separate the creation of the affinity matrix from the t-SNE optimization. Similarly, for the general-purpose methods with subsequent t-SNE, we separate both phases. For the default OT-tSNE affinity construction, we report runtimes using both full SVD and randomized SVD truncated to 200 components (rSVD 200).}
    \label{app:fig:runtime}
\end{figure}

\end{document}